%% file: main.tex
\pdfoutput=1

\documentclass[11pt]{article}

\usepackage[final]{EACL2023}
\usepackage{times}
\usepackage{latexsym}

\usepackage{microtype}

\usepackage[utf8]{inputenc} 
\usepackage[T1]{fontenc}    
\usepackage{hyperref}       
\usepackage{url}            
\usepackage{booktabs}       
\usepackage{amsfonts}       
\usepackage{nicefrac}       
\usepackage{microtype}      
\usepackage{xcolor}         
\usepackage{amssymb}

\usepackage{caption}
\usepackage{subfig}

\usepackage{graphicx}
\usepackage{amsmath}
\usepackage{enumitem}
\usepackage{microtype}

\usepackage{multirow}
\usepackage{multicol}
\usepackage{makecell}
\usepackage{rotating}

\usepackage{pifont}

\setlist[itemize]{noitemsep, topsep=0pt}
\setlist[enumerate]{noitemsep, topsep=0pt}

\newcommand{\xmark}{\ding{55}}
\newcommand{\ernie}[0]{\includegraphics[width=.02\textwidth]{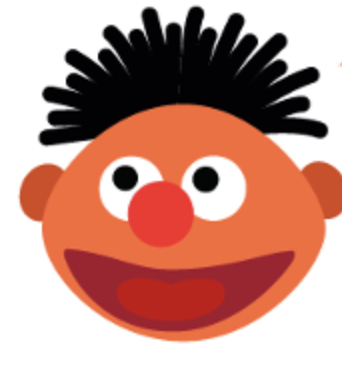}} 
\newcommand{\oscar}[0]{\includegraphics[width=.02\textwidth]{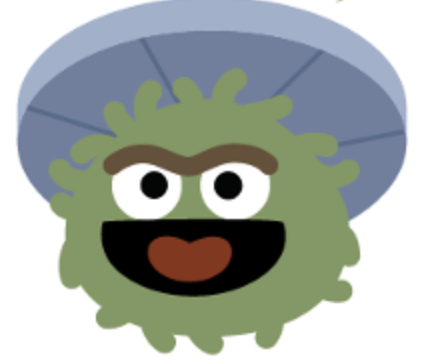}} 
\newcommand{\smileman}[0]{\includegraphics[width=.02\textwidth]{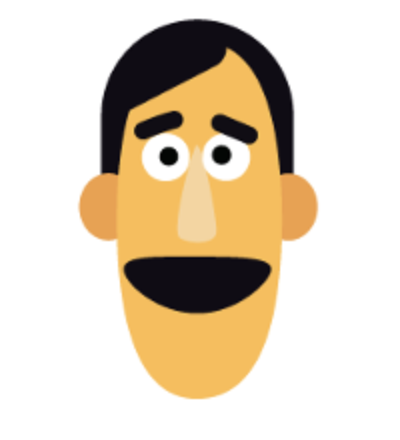}} 
\newcommand{\bday}[0]{\includegraphics[width=.02\textwidth]{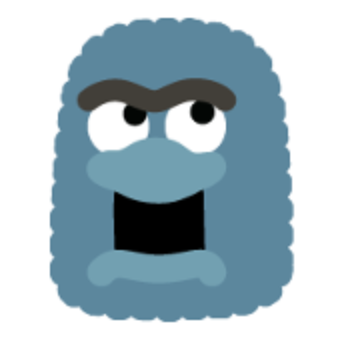}} 
\newcommand{\bippadotta}[0]{\includegraphics[width=.02\textwidth]{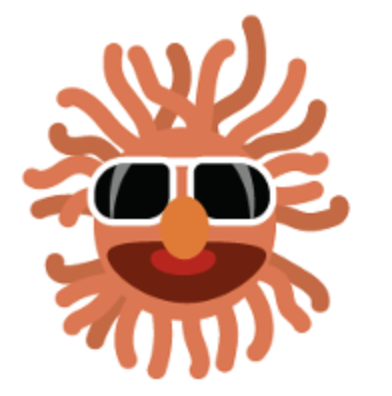}}
\newcommand{\mumford}[0]{\includegraphics[width=.02\textwidth]{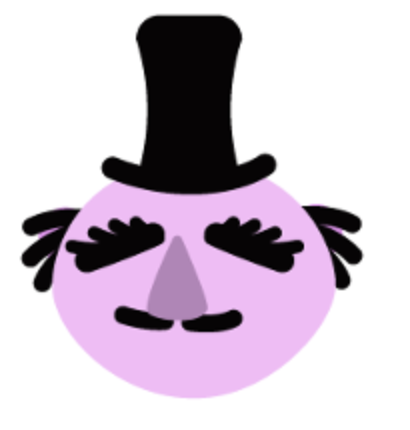}}
\newcommand{\ovejita}[0]{\includegraphics[width=.02\textwidth]{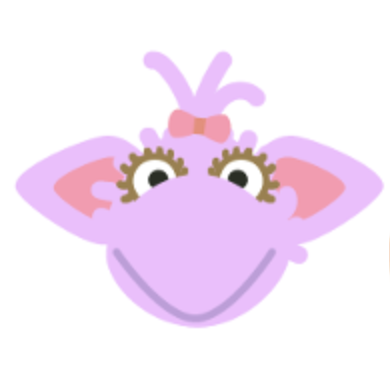}} 
\newcommand{\slimey}[0]{\includegraphics[width=.02\textwidth]{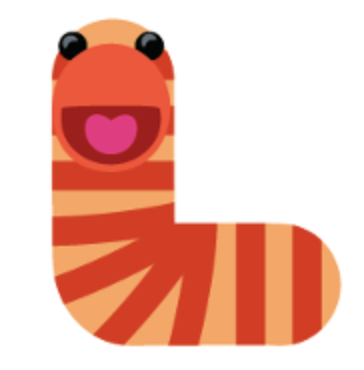}} 
\newcommand{\betty}[0]{\includegraphics[width=.02\textwidth]{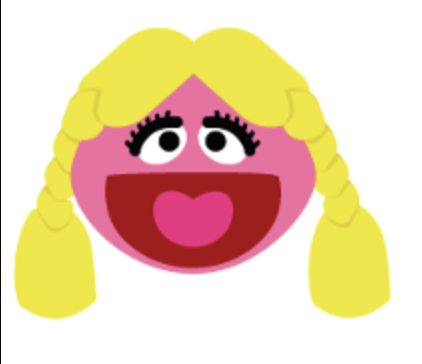}} 
\newcommand{\johnson}[0]{\includegraphics[width=.02\textwidth]{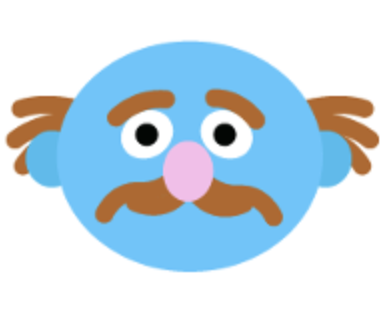}}
\newcommand{\bert}[0]{\includegraphics[width=.02\textwidth]{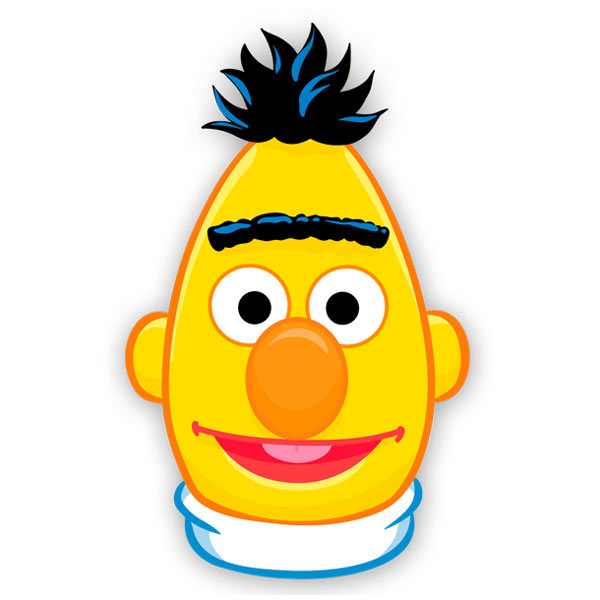}}
\newcommand{\frazzle}[0]{\includegraphics[width=.02\textwidth]{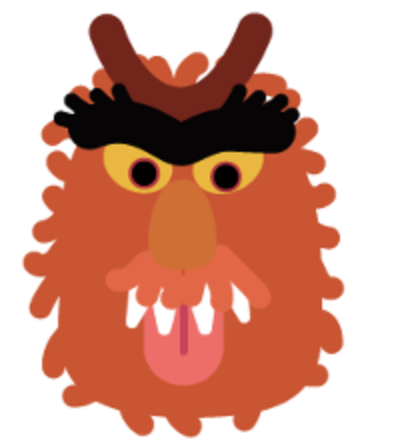}}
\newcommand{\prairie}[0]{\includegraphics[width=.02\textwidth]{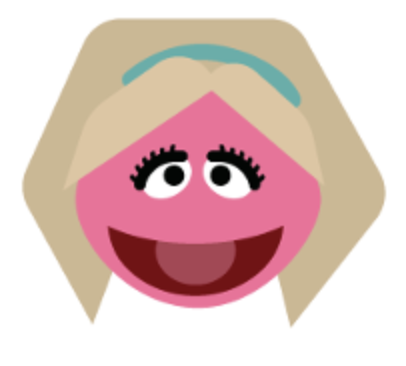}}
\newcommand{\sam}[0]{\includegraphics[width=.02\textwidth]{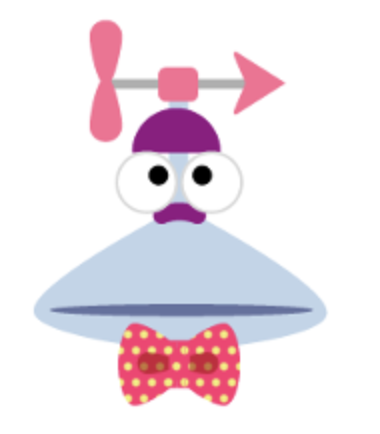}}
\newcommand{\barkley}[0]{\includegraphics[width=.02\textwidth]{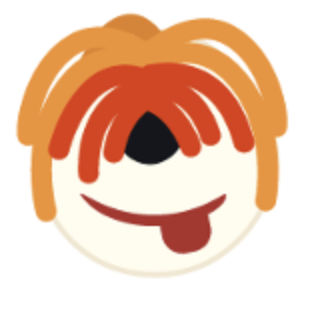}}

\newcommand{\humanpersona}[0]{\includegraphics[width=.02\textwidth]{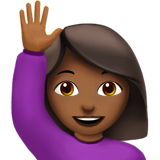}} 
\newcommand{\humanpersoni}[0]{\includegraphics[width=.02\textwidth]{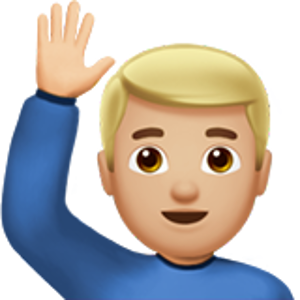}} 

\newcommand{\humanpersonj}[0]{\includegraphics[width=.02\textwidth]{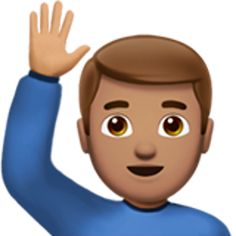}}
\newcommand{\sherlockvalue}[0]{\includegraphics[width=.02\textwidth]{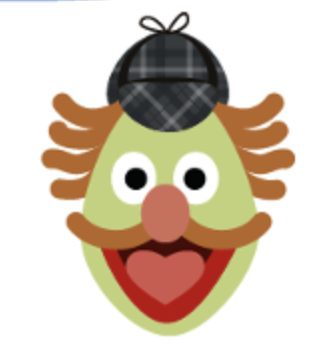}}

\newcommand{\human}[0]{\includegraphics[width=.055\textwidth]{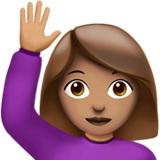}} 
\newcommand{\sherlock}[0]{\includegraphics[width=.055\textwidth]{sesame_street/sherlock.png}}
\newcommand{\erniewinner}[0]{\includegraphics[width=.055\textwidth]{sesame_street/ernie.png}} 
\newcommand{\oscarwinner}[0]{\includegraphics[width=.055\textwidth]{sesame_street/oscar.png}} 
\newcommand{\johnsonwinner}[0]{\includegraphics[width=.055\textwidth]{sesame_street/mrjohnson.png}}

\usepackage{amssymb}

\def\and{\ }

\author{
    ~\textbf{Mark Rofin\textsuperscript{1}}\thanks{\ \ Equal contribution.}, 
    ~\textbf{Vladislav Mikhailov\textsuperscript{1,5}$^*$},
    ~\textbf{Mikhail Florinskiy\textsuperscript{1}$^*$}, \\
    ~\textbf{Andrey Kravchenko\textsuperscript{2}}, 
    ~\textbf{Elena Tutubalina\textsuperscript{1,3}},
    ~\textbf{Tatiana Shavrina\textsuperscript{3,4,5}}, \\
    ~\textbf{Daniel Karabekyan\textsuperscript{1}},
    ~\textbf{Ekaterina Artemova\textsuperscript{6}\thanks{\ \ Work done while at HSE University.}} \\ \\
    \textsuperscript{1}HSE University,
    \textsuperscript{2}University of Oxford\\
    \textsuperscript{3}Artificial Intelligence Research Institute,
    \textsuperscript{4}Institute of Linguistics, RAS, 
    \textsuperscript{5}SaluteDevices \\
    \textsuperscript{6}Center for Information and Language Processing, LMU Munich \\
    \small{
    \textbf{Correspondence:} \href{mailto:Ekaterina.Artemova@lmu.de}{Ekaterina.Artemova@lmu.de}}
}

\title{\textsc{Vote'n'Rank}: Revision of Benchmarking with Social Choice Theory}

\begin{document}

\maketitle

\begin{abstract}
\input{parts/0_abstract}
\end{abstract}

\section{Introduction} \label{sec:intro}
\input{parts/1_intro}

\section{\textsc{Vote'n'Rank}} \label{sec:background}

\subsection{Background} \label{subsec:background}
\input{parts/3_1_background}

\subsection{Aggregation Procedures} \label{subsec:aggregation_procedures}
\input{parts/3_2_procedures}

\subsection{Framework} \label{subsec:framework}
\input{parts/3_3_framework}

\section{Case Studies} \label{sec:case_studies}
\input{parts/4_0_cs}

\subsection{Re-interpreting Benchmarks} \label{subsection:reranking-benchmarks}
\input{parts/4_1_cs1}

\subsection{The Condorcet Winner}
\label{subsec:condorcet_winner}
\input{parts/4_2_cs2}

\subsection{Robustness to Missing Scores}
\label{subsec:missing_scores}

\input{parts/4_3_cs3}

\subsection{Ranking Based on User Preferences} \label{subsec:multifaceted}
\input{parts/4_4_cs4}

\section{Recommendations for Rules Choice}
\input{parts/recommendations}

\section{Conclusion and Future Work}
\input{parts/6_conclusion_fw}

\section{Limitations}
\input{parts/5_limitations}

\section{Ethical Considerations}
\label{appendix:statement}
\input{parts/appendix_statement}

\section*{Acknowledgements}
\input{parts/acknowledgement.tex}

\bibliography{anthology,custom}
\bibliographystyle{acl_natbib}

\clearpage
\newpage


\clearpage
\newpage

\appendix

\section{Aggregation Procedures}
\label{appendix:examples}
\input{parts/appendix_example_rules}

\section{Case Studies}
\label{appendix:case_studies}
\input{parts/appendix_case_studies}


\end{document}

%% file: parts/0_abstract.tex
The development of state-of-the-art systems in different applied areas of machine learning (ML) is driven by benchmarks, which have shaped the paradigm of evaluating generalisation capabilities from multiple perspectives. Although the paradigm is shifting towards more fine-grained evaluation across diverse tasks, the delicate question of how to aggregate the performances has received particular interest in the community. In general, benchmarks follow the unspoken \emph{utilitarian} principles, where the systems are ranked based on their mean average score over task-specific metrics. Such aggregation procedure has been viewed as a sub-optimal evaluation protocol, which may have created the illusion of progress. This paper proposes \textsc{Vote'n'Rank}, a framework for ranking systems in multi-task benchmarks under the principles of the social choice theory. We demonstrate that our approach can be efficiently utilised to draw new insights on benchmarking in several ML sub-fields and identify the best-performing systems in research and development case studies. The \textsc{Vote'n'Rank}'s procedures are more robust than the mean average whilst being able to handle missing performance scores and specify conditions under which the system becomes the winner.

%% file: parts/1_intro.tex
Benchmarking has evolved as a conventional practice for accelerating the development of generalisable systems in different applied areas of machine learning (ML). Benchmarks are typically designed as a collection of datasets, corresponding task-specific evaluation metrics, and a criterion for summarising the overall performance on the tasks~\citep{ruder2021benchmarking}. The benchmark holders provide public leaderboards, which are utilised by ML researchers and practitioners for comparing novel systems against one another, and, if applicable, human baselines, as well as selecting the best-performing ones for practical purposes. According to the benchmark sharing platform \textsc{PapersWithCode}\footnote{\textbf{URL:} \href{https://paperswithcode.com/sota}{\texttt{paperswithcode.com/sota}}. \textbf{Access date}: February 6, 2023.}, the community has put much effort into creating more than $10,000$ influential benchmarks in natural language processing (NLP), computer vision, and knowledge graphs, to name a few.



\vspace{0.7em} \noindent \textbf{Criticism of the benchmark pillars.} The benchmark methodological foundations have received wide criticism from academic and industrial communities~\citep{bowman-dahl-2021-will}. The criticism covers various aspects of benchmarking, raising concerns about the construct validity~\citep{raji2021ai}, fragility of the design and task choices~\citep{dehghani2021benchmark}, data leakage and annotation artifacts~\citep{elangovan-etal-2021-memorization}, SoTA-chasing tendencies at the cost of large carbon footprints~\citep{bender2021dangers}, and low reproducibility of the reported results~\citep{belz-etal-2021-systematic}, inter alia. Recommendations proposed in these studies are of utmost importance to benchmark holders, system users, and developers. However, little attention has been paid to a more nuanced methodological question: \emph{how to aggregate performance scores in multi-task benchmarks?}

\vspace{0.7em} \noindent \textbf{Limits of canonical aggregation.} The appropriateness of mean aggregation in multi-task ML problems is an ongoing debate in the community. The mean aggregation procedure implies that all task metrics are homogeneous~\citep{colombo2022infolm}. Otherwise, it is recommended to evaluate the statistical significance of differences between models with non-parametric tests~\citep{demvsar2006statistical,JMLR:v17:benavoli16a}. In practice, the NLP \textsc{GLUE}-style benchmarks~\citep{wang-etal-2018-glue,wang2019superglue,advglue,liang-etal-2020-xglue} use arithmetic average to rank models over heterogeneous metrics, which
may lead to biased evaluation and subjective outcomes~\citep{niessl2022over,waseem2021disembodied}. The top-leading systems may dominate the others only on the outlier tasks~\citep{agarwal2021deep}, and their ranking is inconsistent with other Pythagorean means~\citep{shavrina2021not}. At the same time, the mean aggregation ignores the relative ordering and relies on the absolute score difference~\citep{peyrard-etal-2017-learning}, equally treating tasks of different complexity~\citep{mishra2021robust} and from different domains~\citep{webb2000multiboosting}.

\vspace{0.7em} \noindent \textbf{Novel aggregation principles.} Recent research has addressed these limitations, introducing novel aggregation methods and principles. One of the directions frames benchmarking in terms of microeconomics, highlighting the importance of the user utility~\citep{ethayarajh-jurafsky-2020-utility}. The other studies urge evaluation of technical system properties in real-world scenarios~\citep{zhou-etal-2021-hulk,dynaboard} and reliability of system rankings~\citep{rodriguez-etal-2021-evaluation}. The benchmarking paradigm is also shifting towards adopting evaluation principles from other fields, such as non-parametric statistics and social choice theory~\citep{choudhury2021linguistically,pmlr-v133-min21a,varshney-etal-2022-ildae,colombobest}.

\vspace{0.7em} \noindent \textbf{Contributions.} Drawing inspiration from the social choice theory, we make two application-oriented contributions and introduce an alternative tool for benchmark evaluation. First, this paper proposes \textsc{Vote'n'Rank}, a flexible framework to rank systems in multi-task/multi-criteria benchmarks and aggregate the performances based on end-user preferences. \textsc{Vote'n'Rank} includes $8$ aggregation procedures that rely on rankings in each criterion and allow to aggregate homogeneous and heterogeneous information. The framework is easy-to-use and allows the users to plug in their own data. Second, we analyse the framework's application in four case studies: \emph{(i)} re-ranking three NLP and multimodal benchmarks; \emph{(ii)} exploring under which circumstances a system becomes \emph{a Condorcet winner}; \emph{(iii)} evaluating robustness to omitted task scores; and \emph{(iv)} ranking systems in accordance with user preferences.

We  publicly release the \textsc{Vote'n'Rank} framework\footnote{\href{https://github.com/PragmaticsLab/vote\_and\_rank}{\texttt{github.com/PragmaticsLab/vote\_and\_rank}}}  to foster further development of reliable and interpretable benchmark evaluation practices for both academic and industrial communities. 

%% file: parts/3_1_background.tex
The study of how individual preferences can be combined to reach a collective decision is the focus of \emph{social choice theory}~\citep{arrow2012social}. There are two main approaches to deal with preferences: \emph{utilitarian} and \emph{ordinal}.
The first approach relies on the so-called cardinal utility, which implies that there exists some unique utility function for each individual that defines their preferences. Here, we can work with utilities as numerical values, and collective decision making aims to maximise the \emph{social welfare utility}. Examples of such utilities are \emph{utilitarian and egalitarian social welfare} measures, where the sum of utilities of individual agents and the utility of the worst agent get maximised, respectively.

The \emph{utilitarian} approach has its drawbacks. First, it implies that kind of utility exists, which is not always true: individuals can compare two systems and prefer one to another but cannot say how many ``utils'' they got. Second, it assumes that individual utilities can be compared. The latter is a solid requirement for benchmarking problems, e.g. when we need to aggregate heterogeneous criteria such as performance and computational efficiency. In order to sum them up, one needs a transformation function that puts the metrics in the same measurement scheme. For example, \textsc{Dynascore}~\citep{dynaboard} utilises Marginal Rate of Substitution (MRS) from economics as such transformation function. Third, the utilitarian compensatory principle is questionable. Can low performance in one task/criterion be compensated by high performance in the others? \citep{munda}

The \emph{ordinal} approach has a weaker requirement, where individuals have preferences ($x$ is preferred to $y$, $x \succ y$, i.e. binary relations over objects), which should be aggregated in social preference (also called social rankings). This approach allows us to aggregate rankings from different tasks and criteria without worrying about measurement schemes. 


%% file: parts/3_2_procedures.tex
\textbf{Definitions.} We adopt the conceptual definitions from the social choice theory to the objectives of selecting the best-performing system and ranking a set of systems as follows: {\it (i) a voter} or a {\it criterion} is a task in a given benchmark, and {\it (ii) an alternative} is a system candidate. 

\vspace{0.7em} \noindent \textbf{Objectives.} Suppose we have a set $M$ of systems $m \in \{m_1,\ldots, m_{|M|}\}$ from the benchmark including a set $T$ of voters $t \in \{t_1,\ldots,t_{|T|}\}$ and the corresponding criteria $S = \{ s_{mt} \}_{m = 1, t = 1}^{m = |M|, t = |T|} $, where $s_{mt}$ is the score of system $m$ in task $t$. Given that, we pursue two main objectives of the aggregation procedure  $\sigma$, $\sigma: S \mapsto (M,\succ_{\sigma})$: {\it (i)} to select the best performing alternative $m^{*}$, so that there is no alternative $\hat{m}, \hat{m} \succ_{\sigma} m^{*}$, and {\it (ii)} to rank the alternatives in the descending order according to $\sigma$ values, so that $m_i \succ_{\sigma} m_j$. Here $\succ_{\sigma}$ denotes the preference resulting from the aggregation procedure $\sigma$.

\vspace{0.7em} \noindent \textbf{Procedures.} We propose $8$ rules from $3$ different classes: \textit{scoring rules}, \textit{iterative scoring rules}, and \textit{majority-relation based rules}. We provide more details and examples in Appendix~\ref{appendix_a:examples}.

\subsubsection{Scoring rules}
The total score of each system is calculated as the sum of corresponding scores in each task $Sc(m)=\sum_{i=1}^{|M|}{c_i p_i(m)}$, where $p_i(m)$ is the number of tasks having model $m$ in the $i^{th}$ place, and $c_i$ is the $i^{th}$ element of the scoring vector $c$. The systems with the highest scores constitute the final decision. We study the following rules that differ in their scoring vectors.
\begin{itemize}
    \item \textit{Plurality rule} applies  $c=(1,0,\ldots,0)$.
    \item \textit{Borda rule} operates on $c=(|M|-1,|M|-2,\ldots,1,0)$.
    \item \textit{Dowdall rule} applies the scoring vector $c=(1, 1/2, \ldots, 1/|M|)$.
\end{itemize}

The scoring vectors are designed to satisfy the voting rules’ properties mentioned in \autoref{tab:properties} in Appendix~\ref{appendix_a:properties}. The scoring vectors’ design is based on the mathematical foundations of the social choice theory and is generally accepted in the community \cite{auizerman1995theory}. 

\vspace{0.7em} \noindent \textbf{Interpretation.} The \textit{Plurality} rule is one of the most widely used in everyday life. It only requires information about the best alternative for each voter. The \textit{Borda} rule takes into account information about all alternatives. It assumes that differences in positions should be treated the same, whether it is between the first and the second alternatives or the second and the third ones. At the same time, the \textit{Dowdall} rule is in some way in-between \textit{Plurality} and \textit{Borda}. It considers information about all alternatives but gives more weight to the difference in the preferences. A similar approach is used in the Eurovision song contest: they use $c=(12, 10, 8, 7,\ldots, 1)$ making the difference in top positions more important to the outcome.

\subsubsection{Iterative scoring rules}
\begin{itemize}
    \item The \textit{Threshold rule} applies $c=(1,1,..,1,1,0)$. In case of ties scoring vectors $(1,1,...,1,0,0)$, ..., $(1, 0,...,0,0)$ are iteratively applied and used only to compare systems with the maximum sum of scores. 
    \item The \textit{Baldwin rule} iteratively applies scoring vectors $(|M|-1,|M|-2,...,1,0)$, $(|M|-2,|M|-3,...,1,0,0)$ ,..., $(1, 0,...,0,0)$, and at each iteration discards systems with the minimum sum of scores.
\end{itemize}

\noindent Both rules stop the procedure when it is impossible to break ties or there is only one alternative left.


\vspace{0.7em} \noindent \textbf{Interpretation.} The rules are similar in their iterative nature but different in terms of the intuition behind them. The \textit{Threshold} rule is based on the idea that the worst position is what matters the most. When we start with $c=(1,1,..,1,1,0)$, we choose the alternatives declared worst in the least amount of cases. Since there can be ties, additional iterations are used to break them with $c=(1,1,...,1,0,0)$ and so on; in other words, by looking at the least-$k$ positions until we have one alternative left or can not break ties.

The \textit{Baldwin} rule has two main differences from \textit{Threshold}. First, it is based on the \textit{Borda} score and considers information from all positions in the ranking, not only the worst one. Second, whilst the \textit{Threshold} rule applies a new vector to the original profile and compares only tied alternatives, the \textit{Baldwin} rule iteratively eliminates the least scored systems and moves the remaining up in rankings. For example, if system $m_A$ is in the fifth place, but alternatives from the first four places are eliminated in the first rounds, $m_A$ will be the first until it is eliminated or is among alternatives in the outcome. 

\subsubsection{Majority-relation based rules}
Let us define a majority relation $\mu$ over the set of alternatives as the following binary relation: $m_A \mu m_B$ iff $m_A$ is ranked higher than $m_B$ by more criteria.
    \begin{itemize}

        \item \textit{Condorcet rule}. $m_C$ is the Condorcet winner (CW) iff $m_C \mu m$ for any $m \in M$.         
        \item \textit{Copeland rule}. Define the lower counter set of systems $m_A$ as a set of systems dominated by $m_A$ via $\mu$: $L(m_A)=\{m \in M, m_A \mu m\}$. In a similar way, define the upper counter set of systems $m_A$ as a set of systems that dominate $m_A$ via $\mu$: $U(m_A)=\{m\in M, m \mu m_A\} $. Define $u(m)=|L(m)|-|U(m)|$. The final decision is provided by the alternatives with the highest $u(m)$.
        \item \textit{Minimax rule}. Let $s(m_A, m_B)$ be the number of criteria for which system $m_A$ is ranked higher than system $m_B$ if $m_A \mu m_B$ or $s(m_A, m_B)=0$ otherwise. The systems are ranked according to the formula $\mathrm{rank}(m_A) = -\max_{B} s(m_B, m_A)$.
    \end{itemize}

\vspace{0.7em} \noindent \textbf{Interpretation.} CW is the alternative that beats all the others in pairwise comparison. However, the \emph{Condorcet} rule does not declare any winner if the CW does not exist. The \emph{Copeland} and \textit{Minimax} rules select the CW whenever it exists and solve the drawback as follows. The \emph{Copeland rule} selects an alternative that dominates more alternatives and is dominated by less (the difference between the numbers is maximised). The \textit{Minimax rule} chooses the alternative with the minimum number of defeats.

\subsection{Properties of the Aggregation Procedures}
\label{subsection:properties}
There is a multitude of voting rules in the social choice theory~\cite{nurmi1983voting,Levin-Nalebuff,de2019systems,aleskerov2010threshold}.  
The motivation behind our rules\footnote{We do not consider more complex rules like \emph{Kemeny} since it is NP-hard to find the Kemeny winner~\cite{bartholdi1989voting}, and it is often implemented as the \emph{Borda} rule approximation~\cite{colombobest}.} is that they generally overcome the mean aggregation limitations and vary in their \textit{properties}, allowing the user to be more flexible in choosing the rule for their purposes. The outcomes can be interpreted in terms of the properties followed or violated by the rules. We discuss our rules' properties in Appendix~\ref{appendix_a:properties}.

%% file: parts/3_3_framework.tex
\begin{figure}[t!]
    \centering
    \scalebox{0.43}{
    \includegraphics[width=\textwidth]{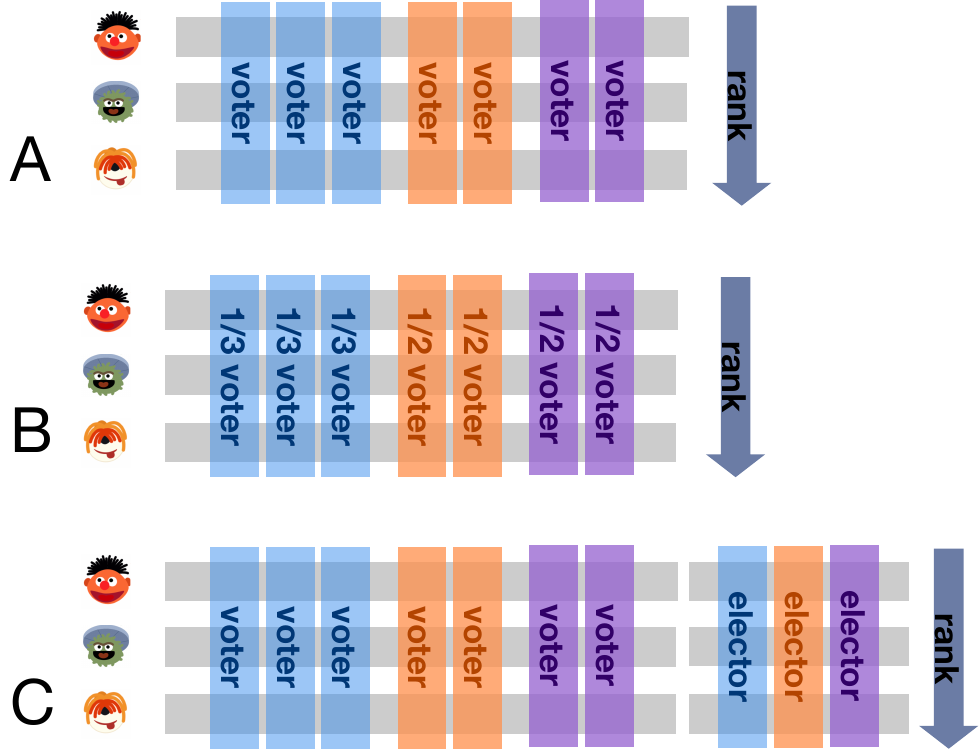}
    }
    \caption{Three ways to run the aggregation procedures. A: Basic aggregation. B: Weighted aggregation. \linebreak C: Two-step aggregation.}\label{fig:ex_leaderboard}
\end{figure}

\autoref{fig:ex_leaderboard} describes three supported settings of performing the aggregation objectives. The toy benchmark has three evaluated alternatives and consists of seven voters grouped by the task, e.g. natural language inference, text classification, and question answering (QA).

\begin{enumerate}[label=\Alph*]
\item \underline{Basic aggregation}: the aggregation procedure is applied to the leaderboard as is.
\item \underline{Weighted aggregation}: each voter in the group is assigned a group weight equal to $1/|T_{group}|$. The blue group weights are $1/3$, and the orange and the violet group weights are $1/2$. Each group contributes equally to the final ranking, regardless of the number of voters. 
\item \underline{Two-step aggregation}: each voter group is treated as a standalone leaderboard. We independently apply a procedure to each voter group and compute an interim ranking shown as ``elector''. Next, we aggregate the group-wise rankings by applying the same procedure one more time and compute the final ranking.
\end{enumerate}

%% file: parts/4_0_cs.tex
This section describes four case studies on three NLP and multimodal benchmarks. Our main objective here is to re-interpret the benchmarking trends under the social choice theory. We provide a brief description of the benchmarks below.


\begin{itemize}[leftmargin=1.em,noitemsep]
    \item \textsc{GLUE} (General Language Understanding Evaluation; \citealp{wang-etal-2018-glue}) combines nine datasets on QA, sentiment analysis, and textual entailment. \textsc{GLUE} also includes a linguistic diagnostic test set. $|M|$=$30$.
    \item \textsc{SGLUE}~\citep{wang2019superglue} is the \textsc{GLUE} follow-up consisting of two diagnostic and eight more complex NLU tasks, ranging from causal reasoning to multi-hop and cloze-style QA. $|M|$=$22$.
    \item \textsc{VALUE} (Video-and-Language Understanding Evaluation;~\citealp{li1value}) covers 11 video-and-language datasets on text-to-video retrieval, video QA, and video captioning. $|M|$=$7$.
\end{itemize}

The leaderboards present the results of evaluating various neural models, such as \textsc{BERT}~\citep{devlin-etal-2019-bert}, \textsc{StructBERT}~\citep{wang2019structbert}, \textsc{ALBERT}~\citep{lan2019albert}, \textsc{RoBERTa}~\citep{liu2019roberta}, \textsc{T5}~\citep{raffel2020exploring}, \textsc{DeBERTa}~\citep{he2020deberta}, \textsc{ERNIE}~\citep{zhang-etal-2019-ernie}, and their ensembles and other model configurations.

%% file: parts/4_1_cs1.tex
\begin{table}[tp!]
    \resizebox{\columnwidth}{!}{
     \input{tables/cs1_ar_kendall}
     }
     \renewcommand\thetable{1}
    \caption{Agreement rates between the top/least-$k$ rankings with $\sigma^{am}$. The Kendall Tau correlation ($\tau$) is computed on the total rankings.} \label{tab:ar_kendall}
\end{table}

\vspace{0.07em} \noindent \textbf{Method.} We begin with a case study on re-ranking systems on the publicly available leaderboards using the scoring and majority-relation based rules\footnote{We omit the iterative rules here for the sake of space.}: \textit{Plurality}, \textit{Dowdall}, \textit{Borda}, \textit{Copeland}, and \textit{Minimax} as the baselines. $\sigma^{og}$ is an aggregation metric that identifies the amount by which the system fails to get a minimum score of $\gamma=0.95$ (lower is better). The comparison is run by computing {\it (i)} the agreement rate (AR; in \%), i.e. the proportion of the top/least-$k$ systems between the given procedure and $\sigma^{am}$, {\it (ii)} the Kendall Tau correlation ($\tau$) between the total rankings, {\it (iii)} the discriminative power (DP) or the number of tied alternatives. i.e. alternatives with the same score~\citep{brandt2016discriminative}, and {\it (iv)} the independence of irrelevant alternatives (IIA), i.e. how often the \emph{new} systems change the ranking (see Appendix~\ref{appendix_a:properties} for details). IIA is computed iteratively in two steps. First, we initialise a leaderboard with two random systems $m_A$ and $m_B$. Second, we add a new random system $m_C$ to the leaderboard and check if the rankings of $m_A$ and $m_B$ have changed. We repeat the procedure by adding up to $|M|$ systems and counting how often the new system affects the ranking. The experiment is run $50$ times to account for randomness. 

\begin{table}[htp!]
    \small
    \centering
    \input{tables/cs1_dp_iia}
    \renewcommand\thetable{2}
    \caption{Discriminative power (DP) and independence of irrelevant alternatives (IIA) values. The lower, the better for both DP and IIA.}  \label{tab:dp_iia}
\end{table}

\vspace{0.08em} \noindent \textbf{Results.}~\autoref{tab:ar_kendall} and~\autoref{tab:dp_iia} present the results except for the \textsc{VALUE} benchmark which is discussed in~\autoref{appendix:case_studies}. We find that methods tend to agree on the top systems, but \textit{Minimax} and \textit{Plurality} disagree on which ones are the worst. Despite high ARs on particular top/least-$k$ systems, the order of the systems on \textsc{GLUE} and \textsc{SGLUE} is different, which is indicated by the low correlation coefficients. The Pythagorean mean results are consistent with one another on the top-$7$ systems and may lead to different worst systems. $\sigma^{og}$ generally disagrees with $\sigma^{am}$ for the top and worst systems on \textsc{GLUE} but has higher ARs and correlation on \textsc{SGLUE}.

At the same time, the DP results demonstrate that \textit{Dowdall} and \textit{Borda} produce only one pair of alternatives with the same score, whilst \textit{Minimax} and \textit{Plurality} treat a significantly larger number of systems as equivalent. The reason is that the rules initially intend to define the best alternative, and they are indecisive between the alternatives when utilised to rank. The IIA experiment shows that introducing a new system influences the \textit{Dowdall} and \textit{Borda} rankings. However, this tendency is less common for \textit{Copeland}, \textit{Minimax}, and \textit{Plurality} and is observed only up to $2$ times on \textsc{SGLUE}.


\input{tables/glue_cs1}

Overall, we observe that the \textsc{GLUE} and \textsc{SGLUE} benchmark rankings depend on the aggregation procedure. The human baseline (\textsc{Human}) rank has risen by up to $13$ positions on \textsc{GLUE} (see~\autoref{tab:glue_cs1}). The \emph{Copeland} method takes \textsc{Human}, \textsc{DeBERTa+CLEVER}, and \textsc{T5} equal, meaning that the difference between the number of candidates they dominate and are dominated by is the same. The \emph{Minimax} ranking suggests that \textsc{Human}, \textsc{T5}, and the \textsc{ALBERT+DAAF+NAS} ensemble are equivalent, meaning that minimal maximum defeats against other models are the same. In their turn, the \textit{Plurality} and \textit{Dowdall} procedures rank \textsc{Human} as the second-best solution, since \textsc{Human} receives the best performance in several tasks, such as \textsc{RTE}~\citep{wang-etal-2018-glue} and \textsc{MNLI}~\citep{williams-etal-2018-broad}. The tendency is also observed on the \textsc{SGLUE} benchmark (see~\autoref{superglue_cs1} in~\autoref{appendix:case_studies}), with the exception that \textsc{Human} is selected as the winner by the \textit{Copeland}, \textit{Plurality}, and \textit{Dowdall} procedures and is equal to the \textsc{ERNIE} system according to \textit{Minimax}. The results for \textit{Borda} are similar to $\sigma^{am}$ and $\sigma^{gm}$ on the top-$4$ and top-$6$ ranks on \textsc{GLUE} and \textsc{SGLUE}, respectively.

\vspace{0.08em} \noindent \textbf{Selecting the winner.} Another application of the voting rules includes selecting the winner from the set of alternatives. Here, we also utilise the \textit{Threshold}, \textit{Baldwin}, and \textit{Condorcet} rules. Note that we run the \textsc{VALUE} experiment over missing and non-missing scores since the \textsc{Human} results are presented for only $6$ out of $11$ tasks. 

\vspace{0.08em} \noindent \textbf{Results.} \autoref{tab:winner} presents the results of selecting the winner for each benchmark. \xmark~denotes that {\it (i)} the given method does not support missing values, or {\it (ii)} there is no \textit{Condorcet} winner (CW). We observe that different SoTAs are selected by $2$/$7$/$3$ (on \textsc{GLUE}/\textsc{SGLUE}/\textsc{VALUE}) procedures as opposed to $\sigma^{am}$, $\sigma^{gm}$, and $\sigma^{og}$. The \textit{Threshold} rule selects \textsc{T5+UDG} and  \textsc{StructBERT+CLEVER} as winners because their performance is the worst the least amount of times. The \textit{Baldwin} rule agrees with the \textit{Plurality} and \textit{Minimax} results. When considering \textsc{VALUE} missing scores, we find that \textsc{Human} is declared SoTA by the \textit{Copeland}, \textit{Minimax}, and \textit{Condorcet} procedures. It means that \textsc{Human} beats any other model in pairwise comparison and is declared the CW, whilst significantly outperforming the systems on specific tasks.

\input{tables/winners}

\vspace{0.07em} \noindent \textbf{Case study discussion.} Benchmarks can suffer from \emph{saturation}, which is characterised by surpassing estimates of the human performance followed by stagnation in SoTA improvements~\citep{ott2022mapping}. The NLP community has discussed saturation of the \textsc{GLUE} benchmark over time~\citep{kiela-etal-2021-dynabench,ruder2021benchmarking} and minor performance gains of the upcoming top-leading systems on \textsc{SGLUE}~\citep{rogers2019transformers}. However, the discussion relies on the mean aggregation. Let us take a step away from the \emph{utilitarian} approach. We observe that \textsc{Human} may still take leading positions, and system ranking varies on these benchmarks under the social choice theory principles. \textsc{VALUE} demonstrates more stable results in terms of the AR and the system order, which we attribute to its novelty and minor performance differences between the systems. Overall, our rules provide interpretable results and cope with the missing leaderboard values in contrast to the \emph{utilitarian} methods.

%% file: tables/cs1_ar_kendall.tex
\begin{tabular}{ccccccccc}

\toprule
\textbf{Benchmark} & $k$ & $\sigma^{gm}$ & $\sigma^{og}$ & \textbf{Copeland} & \textbf{Minimax} & \textbf{Plurality} & \textbf{Dowdall} & \textbf{Borda} \\ \midrule
\multirow{7}{*}{\textsc{GLUE}}
& top-$1$ & 1.0\ & 1.0\ & 1.0\ &   1.0 & 1.0 & 1.0 & 1.0 \\
& top-$3$ & 1.0 &  0.67\ & 1.0 & 0.67 &   0.67 & 0.67 &1.0\\
& top-$5$ & 1.0 & 0.80 & 0.60 & 0.80 &   0.80 & 0.80 &0.8 \\
& top-$7$ &  1.0 & 0.86 &  0.86 & 0.86 &   0.86 & 0.86 &1.0\\
\cmidrule{2-9}
& least-$5$ & 0.67 & 0.00 & 1.0 & 0.33 &   0.33 &  1.0 &1.0\\
& least-$7$ & 0.86 &  0.71 &1.0 & 0.14 &   0.14 &  1.0 &1.0 \\
\cmidrule{2-9}
& $\tau$ & 0.56 & -0.08 & 0.23 & -0.05 & 0.03 & 0.28 & 0.41 \\ \midrule

\multirow{7}{*}{\textsc{SGLUE}}
& top-$1$ & 1.0& 1.0 &  0.00 & 1.0 &   0.00 & 0.00 &   1.0 \\
& top-$3$ & 1.0& 1.0 &  0.67 & 0.67 &   0.67 & 0.67 &   1.0 \\
& top-$5$ &  1.0 & 1.0 &  1.0 & 1.0 &   0.80 & 1.0 &   1.0 \\
& top-$7$ & 1.0 &  0.86&  0.86 & 0.71 &   0.57 & 0.86 &   0.86  \\
\cmidrule{2-9}
& least-$5$ & 1.0 & 1.0 &  1.0 & 0.33 &   0.33 & 1.0 &1.0 \\
& least-$7$ &  0.86  &  0.86& 0.86 & 0.14 &   0.14 & 0.86 &1.0 \\
\cmidrule{2-9}
& $\tau$ & 0.45 & 0.36 & 0.08 & -0.5 & -0.15 & 0.12 & 0.24 \\

\bottomrule

\end{tabular}

%% file: tables/cs1_dp_iia.tex
\begin{tabular}{lcccc}

\toprule
\multirowcell{2}[-0.5ex][l]{\textbf{Method}} & \multicolumn{2}{c}{\textsc{GLUE}} & \multicolumn{2}{c}{\textsc{SGLUE}} \\ \cmidrule(lr){2-3} \cmidrule(lr){4-5} 
 & DP & IIA & DP & IIA  \\ \midrule
$\sigma^{am}$ & 1\  & 0.0\ \tiny{$\pm$0.0}  & 0\  & 0.0\ \tiny{$\pm$0.0}  \\
$\sigma^{gm}$ & 0\  & 0.0\ \tiny{$\pm$0.0}  & 0\  & 0.0\ \tiny{$\pm$0.0}  \\
$\sigma^{og}$ & 3\  & 0.0\ \tiny{$\pm$0.0}  & 1\  & 0.0\ \tiny{$\pm$0.0} \\
\textbf{Copeland} & 6\  & 2.76 \tiny{$\pm$1.3}  & 2\  & 0.90\ \tiny{$\pm$0.8}  \\
\textbf{Minimax} & 21\  & 2.94 \tiny{$\pm$1.5}\  & 17\  & 1.14\ \tiny{$\pm$1.0}  \\
\textbf{Plurality} & 25\  & 5.26\ \tiny{$\pm$1.6}  & 17\  & 1.98\ \tiny{$\pm$1.4}  \\
\textbf{Dowdall} & 0\  & 9.10\ \tiny{$\pm$2.4}   & 0\  & 4.24\ \tiny{$\pm$2.1}  \\
\textbf{Borda} & 0\  & 7.96\ \tiny{$\pm$3.8}  & 1\  & 5.38\ \tiny{$\pm$1.8} \\

\bottomrule
\end{tabular}

%% file: tables/glue_cs1.tex
\begin{table*}[ht!]
\centering
\resizebox{1\linewidth}{!}{
\small
\begin{tabular}{ccccccccc}
\toprule
\textbf{Rank} & $\sigma^{am}$ & $\sigma^{gm}$ & $\sigma^{og}$ & \textbf{Copeland} & \textbf{Minimax} & \textbf{Plurality} & \textbf{Dowdall} & \textbf{Borda}\\ \midrule

1 & \ernie$^{91.18}$ & \ernie$_{\updownarrow0}^{90.89}$ & \ernie$_{\updownarrow0}^{0.074}$ & \ernie$_{\updownarrow0}^{29.00}$ & \ernie$_{\updownarrow0}^{0}$ &   \ernie$_{\updownarrow0}^{2.00}$ & \ernie$_{\updownarrow0}^{4.95}$ & \ernie$_{\updownarrow0}^{260.50}$ \\

2 & \oscar$^{91.07}$ &\oscar$_{\updownarrow0}^{90.78}$ & \ovejita$_{\uparrow4}^{0.075}$ &\smileman$_{\uparrow1}^{25.00}$ &   \smileman$_{\uparrow1}^{-5.50}$ &\humanpersona$_{\uparrow13}^{2.00}$ &  \humanpersona$_{\uparrow13}^{4.08}$ &\oscar$_{\updownarrow0}^{256.00}$ \\

3 & \smileman$^{90.88}$ &   \smileman$_{\updownarrow0}^{90.56}$ & \oscar $_{\downarrow1}^{0.076}$ &  \oscar$_{\downarrow1}^{24.00}$ &  \ovejita$_{\uparrow1}^{-6.00}$ & \smileman$_{\updownarrow0}^{1.50}$ &   \smileman$_{\updownarrow0}^{3.82}$ &   \smileman$_{\updownarrow0}^{247.50}$ \\

4 & \ovejita$^{90.86}$ &  \ovejita$_{\updownarrow0}^{90.48}$ & \bippadotta$_{\updownarrow0}^{0.076}$ &\slimey$_{\uparrow3}^{22.00}$ &  \oscar$_{\downarrow2}^{-6.50}$ &  \bday$_{\uparrow1}^{1.00}$ &  \ovejita$_{\updownarrow0}^{3.41}$ &  \ovejita$_{\updownarrow0}^{241.50}$ \\

5 & \bday$^{90.74}$ &\bday$_{\updownarrow0}^{90.44}$ & \bday$_{\updownarrow0}^{0.077}$ &  \humanpersona$_{\uparrow10}^{22.00}$ &\slimey$_{\uparrow2}^{-7.00}$ &\oscar$_{\downarrow3}^{1.00}$ &  \oscar$_{\downarrow3}^{3.27}$ &\bippadotta$_{\uparrow1}^{233.50}$ \\

6 & \bippadotta$^{90.66}$ &\bippadotta$_{\updownarrow0}^{90.34}$ & \slimey$_{\uparrow1}^{0.078}$ &  \ovejita$_{\downarrow2}^{22.00}$ &   \humanpersona$_{\uparrow9}^{-7.00}$ &  \bippadotta$_{\updownarrow0}^{0.50}$ &  \bday$_{\downarrow1}^{2.57}$ &\slimey$_{\uparrow1}^{229.50}$ \\

7 & \slimey$^{90.48}$ &\slimey$_{\updownarrow0}^{90.11}$ & \johnson$_{\uparrow3}^{0.082}$ &\bippadotta$_{\downarrow1}^{16.00}$ &  \bippadotta$_{\downarrow1}^{-7.00}$ &  \ovejita$_{\downarrow3}^{0.00}$ &\slimey$_{\updownarrow0}^{2.55}$ &  \bday$_{\downarrow2}^{220.50}$ \\

\bottomrule
\end{tabular}
}
\renewcommand\thetable{3}
\caption{Results of re-ranking the GLUE benchmark. Changes in the system ranks are depicted with arrows, whilst the superscripts denote scores assigned by the aggregation procedure. \underline{Notations}: \humanpersona=\textsc{Human}; \ernie=\textsc{ERNIE}; \oscar=\textsc{StructBERT+CLEVER}; \smileman=\textsc{DeBERTa+CLEVER};\ovejita=\textsc{DeBERTa/TuringNLRv4};
\bday=\textsc{MacALBERT+DKM};\slimey=\textsc{T5};
\bippadotta=\textsc{ALBERT+DAAF+NAS};\johnson=\textsc{Funnel}. The superscript values stand for the voting rules' scores, whilst the subscript values indicate changes in the ranking positions. $\uparrow x$ means up $x$ positions, $\downarrow x$ means down $x$ positions, $\updownarrow$ means no changes.}
\label{tab:glue_cs1}
\end{table*}

%% file: tables/winners.tex
\newcommand{\erniewinnerresize}[0]{\resizebox{.02\textwidth}{!}\erniewinner}
\newcommand{\sherlockresize}[0]{\resizebox{.02\textwidth}{!}\sherlock}
\newcommand{\humanresize}[0]{\resizebox{.02\textwidth}{!}\human}
\newcommand{\johnsonwinnerresize}[0]{\resizebox{.02\textwidth}{!}\johnsonwinner} 
\newcommand{\oscarwinnerresize}[0]{\resizebox{.02\textwidth}{!}\oscarwinner}

\begin{table}[t!]
    \centering
    \small
    \begin{tabular}{lccc}
        \toprule
        \textbf{Method} & \textsc{GLUE} & \textsc{SGLUE} & \textsc{VALUE} \\
        \midrule
        $\sigma^{am}$& \erniewinnerresize & \erniewinnerresize & \xmark/\sherlockresize  \\
        $\sigma^{gm}$ &\erniewinnerresize & \erniewinnerresize & \xmark/\sherlockresize \\
        $\sigma^{og}$ & \erniewinnerresize & \erniewinnerresize & \xmark / \sherlockresize \\
        \textbf{Copeland} & \erniewinnerresize & \humanresize & \humanresize/\sherlockresize \\
        \textbf{Minimax} & \erniewinnerresize & \erniewinnerresize \humanresize & \humanresize/\sherlockresize \\
        \textbf{Plurality} & \erniewinnerresize \humanresize & \humanresize & \xmark/\sherlockresize \\
        \textbf{Dowdall} & \erniewinnerresize & \humanresize & \xmark/\sherlockresize \\
        \textbf{Borda} & \erniewinnerresize & \erniewinnerresize & \xmark/\sherlockresize \\ 
        \textbf{Threshold} & \oscarwinnerresize & \johnsonwinnerresize & \xmark/\sherlockresize \\
        \textbf{Baldwin} & \erniewinnerresize & \erniewinnerresize \humanresize & \xmark/\sherlockresize \\ 
        \textbf{Condorcet} & \erniewinnerresize & \xmark & \humanresize/\sherlockresize \\
        \bottomrule
    \end{tabular}
\caption{The winner selection results. \underline{Notations}: 
    \oscarwinnerresize=\textsc{StructBERT+CLEVER};
    \humanresize=\textsc{Human};
    \sherlockresize=\textsc{craig.starr};
    \erniewinnerresize=\textsc{ERNIE};\johnsonwinnerresize=\textsc{T5+UDG}.}\label{tab:winner}
\end{table}

%% file: parts/4_2_cs2.tex
\begin{figure*}[t!]
    \centering
    \subfloat[\textsc{GLUE}]{\includegraphics[width=0.3\textwidth]{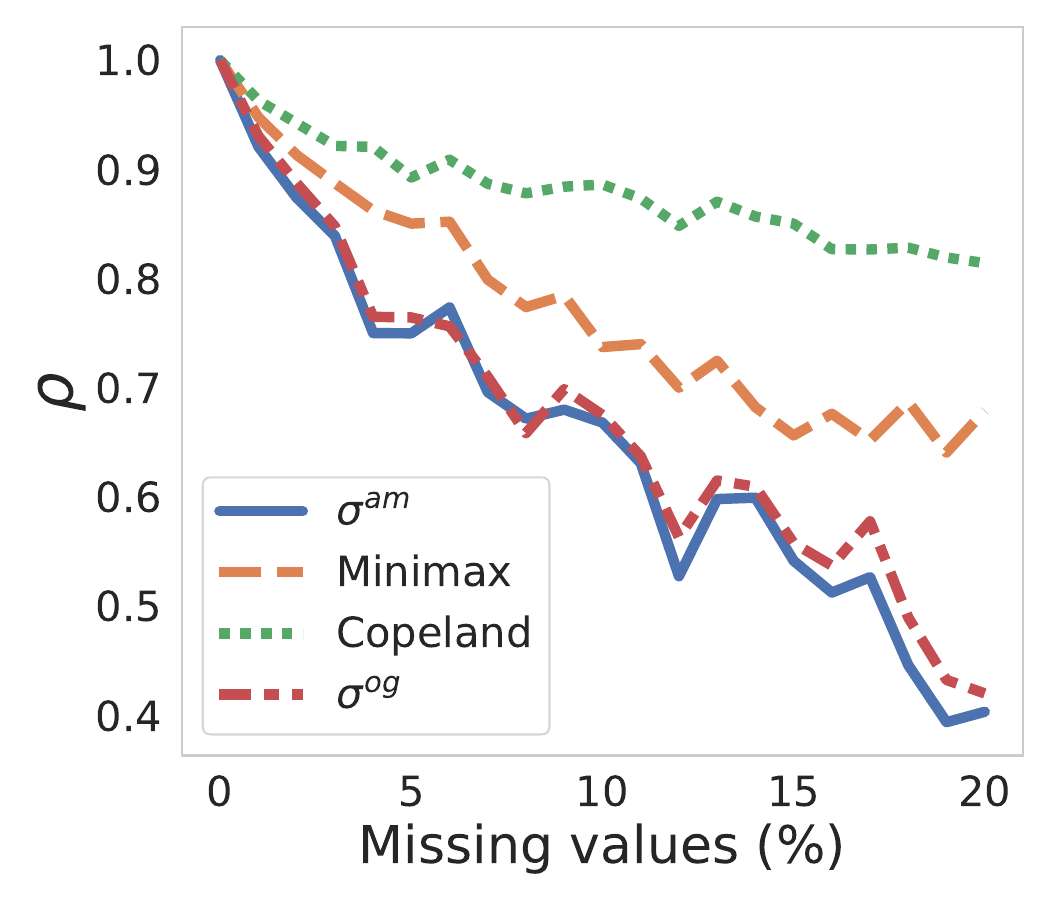}}
    \subfloat[\textsc{SGLUE}]{\includegraphics[width=0.3\textwidth]{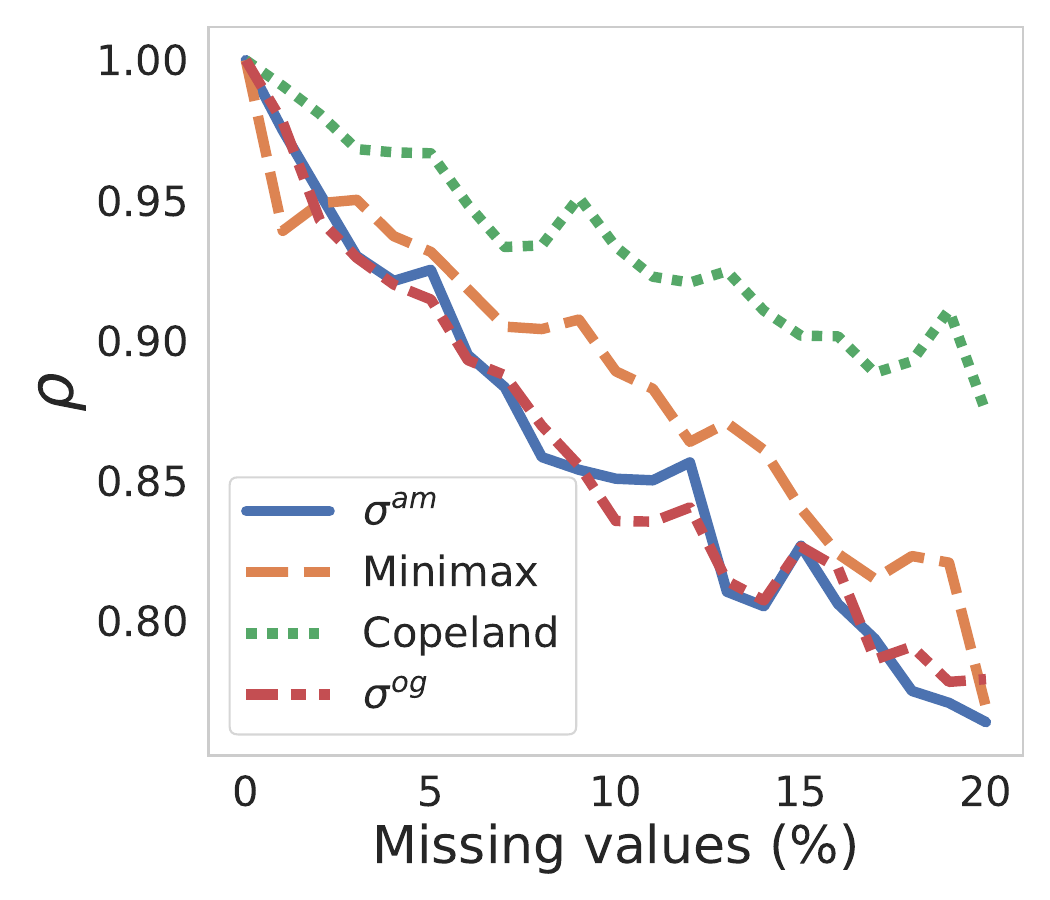}}
    \subfloat[\textsc{VALUE}]{\includegraphics[width=0.3\textwidth]{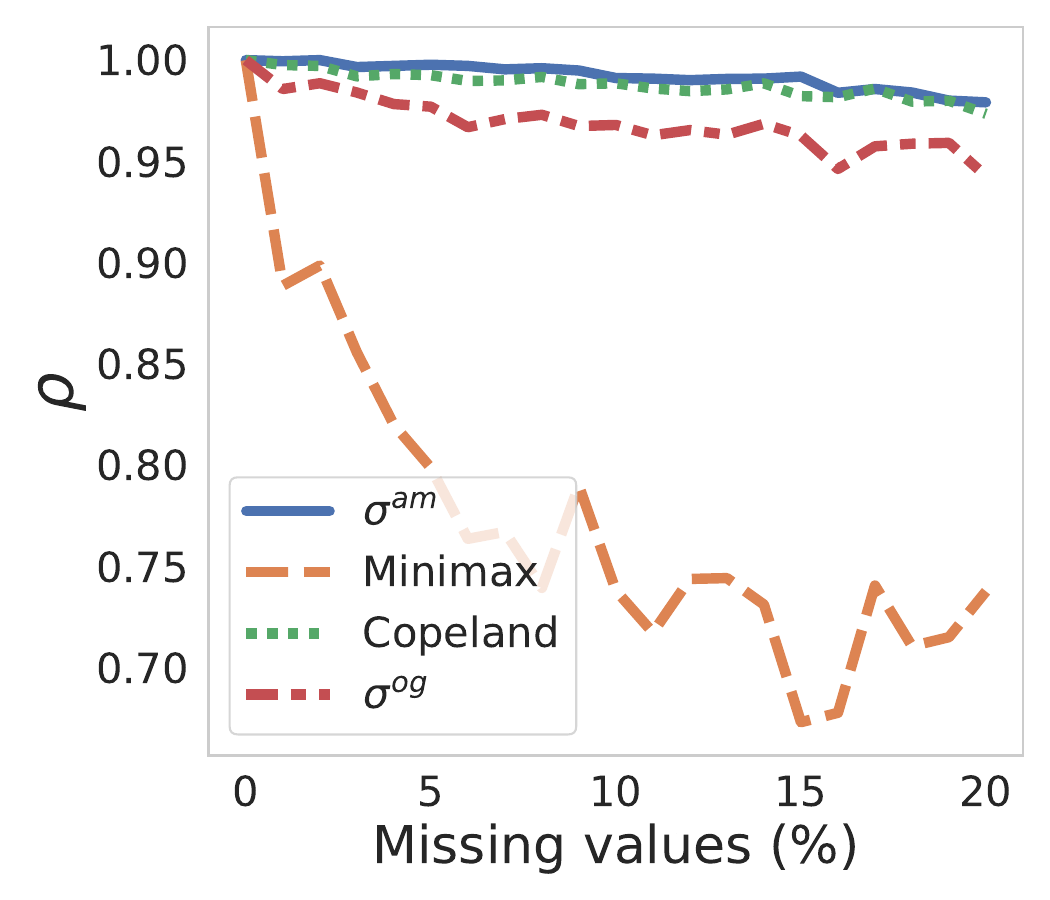}}
    \caption{Spearman correlation ($\rho$) between top-$7$ model rankings with/without omitted leaderboard values for $\sigma^{am}$, $\sigma^{og}$, \textit{Minimax}, and \textit{Copeland} rankings. The results are averaged over $100$ runs.}
    \label{fig:cs3}
\end{figure*}


One of the most natural ways to choose the best system given a set of weights defined by the user is the Condorcet method, which declares a system the winner if it dominates all other alternatives in pairwise comparison~\citep{black1958theory}. The Condorcet method is hard to destabilise~\citep{edelman2015myth} and easy to interpret in practice, indicating that the CW best matches the preferences. Given the weights vector, finding the CW, if it exists, is trivial. We can also find the weights that make a given alternative the CW or determine that no weights with that property exist.


\vspace{0.07em} \noindent\textbf{Method.} Let us define an operator $R(m_1, m_2, i)$:

\begin{equation}
\resizebox{0.42\textwidth}{!}{%
    $R(m_1, m_2, i) = 
    \begin{cases}
    1, & \text{if } \exists s_{m_1 i} \wedge \exists s_{m_2 i} \wedge s_{m_1 i} > s_{m_2 i}, \\
    -1, & \text{if } \exists s_{m_1 i} \wedge \exists s_{m_2 i} \wedge s_{m_1 i} < s_{m_2 i}, \\
    0, & \text{otherwise}
    \end{cases}$
    }
\end{equation}

A system $m$ is declared a CW if the following property is satisfied:

\begin{equation}
\label{eq:condorcet_scalar}
\resizebox{0.4\textwidth}{!}{%
$\forall m^\prime \in M \setminus \{m\} \quad \sum_{k = 1}^{|T|} R(m, m^\prime, k) w_k \geq 0$
}
\end{equation}

Let $G_m \in \{-1, 0, 1\}^{|M| - 1 \times |T|}$ be a matrix such that $G_{ij} = R(m, m_i^\prime, j)$, where $ \{m^\prime_1, .., m^\prime_{|M| - 1} \} = M \setminus \{ m \}$.

\vspace{0.4em} \noindent  \autoref{eq:condorcet_scalar} can be re-written as: $G_m \cdot w \succcurlyeq 0$, which results in defining a space $W^*$ in $\mathbb{R}^{|T|}$, whose each point is a weight vector making $m$ a CW. Any linear algorithm can be applied to find a point in $W^*$ or determine that $W^*$ is empty. Furthermore, any other linear conditions can be added, such as upper/lower bounds of the $w$ components and a linear function that needs to be optimised, e.g. $w_i \longrightarrow \min$. Let us call a system for which there exists a vector of weights making it a CW \textit{prospective}. By definition, the system is \emph{prospective} if $W^*$ is not empty.

\vspace{0.07em} \noindent \textbf{Example.} Let us illustrate the method on the \textsc{SGLUE} benchmark (see~\autoref{tab:winner}). There is no CW if the task weights are assigned uniformly. Nevertheless, \textsc{T5} may become the CW when the \textsc{BoolQ} accuracy~\citep{clark-etal-2019-boolq} and \textsc{MultiRC} exact match scores~\citep{khashabi-etal-2018-looking} have equal weights of $0.5$, and the other criteria weights are zeroed. In this scenario, \textsc{T5} has been found to be a \emph{prospective} system on \textsc{SGLUE}, whiste \textsc{RoBERTa} is declared \emph{non-prospective}.

\vspace{0.07em} \noindent  \textbf{Results.} There are $9$/$88$, $10$/$12$, and $3$/$3$ prospective/non-prospective systems on \textsc{GLUE}, \textsc{SGLUE}, and \textsc{VALUE}, respectively. The results indicate that it is possible to find specific evaluation scenarios in which a given system is the best. In contrast, the non-prospective system always has an alternative that performs neither worse nor better.

\vspace{0.7em} \noindent \textbf{Case study discussion.} The CW criterion presents another perspective of selecting the best systems. Notably, the existence of the CW weights assumes that practitioners can simulate a set of real-world scenarios where the system is the best across the given axes. Specifying if the system can be the CW on the leaderboard would help diagnosing the systems without additional heavy experiments. The developers also can document this information on model sharing platforms, e.g. \textsc{HuggingFace}~\citep{wolf-etal-2020-transformers}.

%% file: parts/4_3_cs3.tex
This case study considers a more detailed analysis of the majority-relation based voting rules that can be efficiently utilised for ranking systems and selecting the winner over missing scores. Here, we evaluate the robustness of the rules to omitted performance scores and analyse how the rankings change under such perturbation.

\vspace{0.7em} \noindent \textbf{Method.} \textit{Copeland} and \textit{Minimax} take as input the majority graph in which each vertex corresponds to a candidate and an edge from the candidate $m_1$ to $m_2$ exists iff $m_1 \mu m_2$, i.e. $m_1$ is ranked higher than $m_2$ by more criteria. Let us say there are $T$ criteria $t \in \{t_1,\ldots,t_T\}$ and $w \in \{w_1,\ldots,w_n\}$ are the weights assigned to them.
\begin{equation}
\resizebox{0.42\textwidth}{!}{%
$m_1 \mu m_2 \iff \sum_{i=1}^{|T|} w_i R(m_1, m_2, i) > 0$
}
\end{equation}

\noindent When evaluating $R(m_1, m_2, i)$, this approach can handle missing values, ignoring the pairs where either of the scores is missing. We can apply the majority-relation based rules using relation $\mu$ to rank alternatives with missing scores without losing any information whilst accounting for the available criteria. 

We analyse the robustness of the \textit{Copeland} and \textit{Minimax} rules as follows. First, we compute the rankings using both methods on each benchmark without omitting scores and use them as references. Next, we randomly replace $N$ scores with empty values and find top-$7$ systems over the corrupted leaderboards. We calculate the Spearman correlation ($\rho$) between the final rankings and the references. Note that we use the median values when omitting scores for $\sigma^{am}$ and $\sigma^{og}$ as the baselines.

\vspace{0.7em} \noindent \textbf{Results.} ~\autoref{fig:cs3} shows that $\sigma^{am}$ and $\sigma^{og}$ display lower stability and \textit{Copeland} performs the best on \textsc{GLUE} and \textsc{SGLUE}. However, we observe that \textit{Minimax} is the least stable on \textsc{VALUE}, whilst \textit{Copeland}, $\sigma^{am}$, and $\sigma^{og}$ perform on par. 

\vspace{0.7em} \noindent \textbf{Case study discussion.} We attribute the low stability of \textit{Minimax} on \textsc{VALUE} to its limitations. Recall that there are minor differences between the systems on \textsc{VALUE}, which cause \textit{Minimax} to score them very similar (see~\autoref{tab:value_cs1} in~\autoref{appendix:case_studies}). In this case any missing value can influence the rankings, which results in the low $\rho$ coefficients.

%% file: parts/4_4_cs4.tex
\input{tables/glue_cs4}

This case study aims at system ranking based on the user utility. We rank systems in a simulated scenario that considers preferences on \emph{performance}, \emph{computational efficiency}, and \emph{fairness}.

\vspace{0.07em} \noindent \textbf{Method.}  We use the \textsc{HuggingFace} library to fine-tune and evaluate systems on \textsc{GLUE}. Each system is initialised with a fixed set of five random seeds and fine-tuned for five epochs with default hyper-parameters and a batch size of $16$. The development set performance is averaged across all runs. We consider the following systems: BERT-base, RoBERTa-base, ALBERT-base, DeBERTa-base, DistilBERT-base~\citep{sanh2019distilbert}, DistilRoBERTa-base~\citep{sanh2019distilbert}, and GPT2-medium~\citep{radford2019language}. The experiments are run on a single GPU unit, NVIDIA A100 80 GB SXM (NVLink), $4$-CPU cores, AMD EPYC 7702 2-3.35 GHz, and 1 TB RAM.

The \emph{efficiency} is computed during fine-tuning via the Impact tracker toolkit~\citep{henderson2020towards}: the total power, run time in hours, GPU usage in hours, and estimated carbon footprint. To maximise these, we inverse the computational efficiency features through multiplying them by $-1$.  

To measure \emph{fairness}, we choose three social bias evaluation datasets: \textsc{CrowS-Pairs}~\citep{nangia-etal-2020-crows}, \textsc{StereoSet}~\citep{nadeem-etal-2021-stereoset}, and \textsc{Winobias}~\citep{zhao-etal-2018-gender}. In these datasets, one sentence is always more stereotyping than the other. Following~\citet{nangia-etal-2020-crows}, we use MLM scoring~\citep{salazar-etal-2020-masked} to score the pairs. The final metrics account for cases (\%) in which a less stereotyping sentence is the most probable. 

For the sake of space, we present the results on the \textit{Borda} procedure in the basic, weighted, and two-step aggregation settings (\S\ref{subsec:framework}). We assign the weights vector as $(0.4, 0.3, 0.3)$ to performance, efficiency, and fairness. The weights are introduced to increase the impact of performance. We use $\sigma^{am}$ as the baseline and interim rankings by each criterion individually as references.

\vspace{0.07em} \noindent \textbf{Results.} \autoref{table:two_step_borda} shows that \textsc{DeBERTa} is the winner according to $\sigma^{am}$ and \textit{Borda}. However, it requires more computational resources than the other systems and is mediocre in detecting social biases. As a result, it is not the best system in any user-oriented ranking. In this scenario, \textit{Borda} tends to favour the distilled systems (\textsc{DistilRoBERTa} and \textsc{DistilBERT}) due to their computational efficiency, which has the highest impact on the ranking with four criteria assigned per task. The weighted \textit{Borda} ranks \textsc{DistilBERT}, \textsc{ALBERT}, and \textsc{BERT} as the top-$3$ systems. In its turn, the weighted $2$-step \textit{Borda} prefers \textsc{ALBERT} first, followed by \textsc{BERT} and \textsc{DistilBERT}. \textsc{ALBERT} is selected as the winner by the fairness ranking only and occupies the middle positions in the two other rankings. \textsc{RoBERTa} drops down drastically from the second rank ($\sigma^{am}$), whilst \textsc{GPT2} remains in the least-$3$ systems.

\vspace{0.07em} \noindent \textbf{Case study discussion.} Overall, our setup follows \textsc{Dynascore}~\citep{dynaboard}, where the microeconomic concept of MRS is used to compare performance, efficiency, and fairness metrics, followed by the weighted average score as the final ranking. Unlike the \textsc{Dynascore} results, we find that the average performance ranking is not preserved when using our voting rules. The most notable difference is with \textsc{DeBERTa} and \textsc{RoBERTa} systems, which may become penalised for low efficiency in our case. The reason is that in \textsc{Dynascore}, the weight of $0.5$ is assigned to performance which blocks substantial changes in re-ranking.

%% file: tables/glue_cs4.tex
\begin{table*}[t!]
\centering
\resizebox{1\linewidth}{!}{
\scriptsize
    \begin{tabular}{cccccccc}
    \toprule
    \textbf{Rank} & $\sigma^{am}_{\text{\textbf{Performance}}} $ & \textbf{Borda} & \makecell{\textbf{Weighted} \\ \textbf{Borda}} & \makecell{\textbf{Weighted} \\ \textbf{2-step Borda}}  & \makecell{\textbf{Borda} \\ \textbf{Performance}} &  \makecell{\textbf{Borda} \\ \textbf{Efficiency}} &  \makecell{\textbf{Borda} \\ \textbf{Fairness}}\\
    \midrule
    1 & \smileman$^{82.73}$ & \ovejita$_{\uparrow4}^{267.0}$  & \sam$_{\uparrow5}^{10.75}$   &   \bday$_{\uparrow2}^{4.30}$          &   \smileman$_{\updownarrow0}^{56.5}$    &   \ovejita$_{\uparrow4}^{223.0}$   &  \bday$_{\uparrow2}^{19.0}$\\  
    2 & \prairie$^{82.52}$  & \sam$_{\uparrow4}^{245.0}$    & \bday$_{\uparrow1}^{ 9.83}$  &   \bert$_{\uparrow2}^{3.60}$          &   \prairie$_{\updownarrow0}^{49.0}$    &   \sam$_{\uparrow4}^{216.0}$    &   \sam$_{\uparrow4}^{18.0}$ \\
    3 & \bday$^{80.94}$     & \bert$_{\uparrow1}^{166.0}$   & \bert$_{\uparrow1}^{8.96}$  &   \sam$_{\uparrow3}^{3.40}$        &   \bday$_{\updownarrow0}^{32.5}$    &   \bert$_{\uparrow1}^{120.0}$   &   \bert $_{\uparrow1}^{14.0}$ \\
    4 & \bert$^{79.20}$     & \bday$_{\downarrow1}^{154.0}$ & \ovejita$_{\uparrow1}^{8.63}$   &   \smileman$_{\downarrow3}^{3.00}$    &   \bert$_{\updownarrow0}^{32.0}$	&   \bday$_{\downarrow1}^{103.0}$	    &   \smileman$_{\downarrow3}^{11.00}$\\
    5 & \ovejita$^{78.56}$    & \prairie$_{\downarrow3}^{144.0}$  & \smileman$_{\downarrow4}^{7.17}$  &   \ovejita$_{\updownarrow0}^{2.90}$ &   \ovejita$_{\updownarrow0}^{17.0}$  &  \prairie$_{\downarrow3}^{91.0}$  &   \barkley$_{\uparrow1}^{11.0}$  \\
    6 & \sam $^{77.89}$     & \barkley$_{\uparrow1}^{10.0}$  &  \prairie$_{\downarrow4}^{7.04}$  &   \prairie$_{\downarrow3}^{2.60}$     &   \sam$_{\updownarrow0}^{11.0}$    &   \barkley$_{\uparrow1}^{84.0}$   &  \ovejita$_{\uparrow1}^{7.00}$ \\
    7 & \barkley$^{75.95}$  & \smileman$_{\downarrow6}^{70.50}$  & \barkley$_{\updownarrow0}^{5.47}$  &   \barkley$_{\updownarrow0}^{0.90}$   &   \barkley$_{\updownarrow0}^{8.0}$     &  \smileman$_{\downarrow6}^{3.00}$  &  \prairie$_{\downarrow5}^{4.00}$  \\
    \bottomrule
\end{tabular}
}
\caption{Results of re-ranking the \textsc{GLUE} benchmark using the \textit{Borda} rule in the simulated user-oriented scenario. \underline{Notations}: \bday = \textsc{ALBERT}; 
\bert=\textsc{BERT}; \sam=\textsc{DistilBERT};
\prairie=\textsc{RoBERTa}; \ovejita=\textsc{DistilRoBERTa};
\smileman=\textsc{DeBERTa};
\barkley=\textsc{GPT2}.}
\label{table:two_step_borda}
\end{table*}

%% file: parts/recommendations.tex
The information about the voting rules' properties can be used to choose the most suitable one to the user's preferences~\cite{felsenthal2012electoral}. We also provide the following recommendations.

\begin{itemize}
    \item The \emph{Plurality} rule is a good choice if the user wants only the best systems in \emph{each} criterion.
    \item If \emph{all} ranking positions matter, use the \emph{Borda} or \emph{Dowdall} rules. Note that \emph{Dowdall} assigns higher weights to the top positions.
    \item The \textit{Threshold} rule is helpful in cases when the user wants to minimise the number of the low-performance criteria: the rule assigns the highest rank to the system that is considered the worst in the least amount of criteria.
    \item If the goal is to select the system that beats all the others in pairwise comparison, use the \textit{Baldwin}, \textit{Condorcet}, \textit{Copeland}, or \textit{Minimax} rules. These rules are Condorcet consistent; i.e. choose the CW if it exists. The main difference is how the rules behave when there is no CW. In particular, \textit{Baldwin} selects the system that is left after elimination according to the \textit{Borda} scores. \textit{Copeland} chooses the system that dominates the others in more cases and is dominated by the least. In turn, \textit{Minimax} selects the system with minimum defeat in pairwise comparison. 
    \item The outcomes \emph{may} contain equivalent alternatives (\S\ref{subsection:reranking-benchmarks}). Depending on the scenario, the user can select the rule that produces ties with a lower probability or \textit{Dowdall} and \textit{Borda} if their properties meet the preferences.
\end{itemize}


%% file: parts/6_conclusion_fw.tex
This paper introduces novel aggregation procedures to rank and select the best-performing systems under the social choice theory principles. Our approach provides an alternative perspective of system evaluation in benchmarking and overcomes the standard mean aggregation limitations. 

Our case studies show that \textsc{Vote'n'Rank} provides interpretable decisions on the best and worst systems whilst accounting for missing performance scores and potential user preferences. The framework allows for finding scenarios in which a given system dominates the others. At the same time, the rule choice may depend on the particular research and development purpose. We provide recommendations based on the rules' properties and scenarios of the intended framework's application.

The application scope of \textsc{Vote'n'Rank} is not limited and may be easily extended to other applied ML areas. The current mainstay of \emph{multilingual} and \emph{multimodal} benchmarking fails to provide a nuanced comparison of systems across languages and tasks. In our future work we hope to explore applications of the social choice theory in this direction through a consideration of user studies and an extended set of voting rules and linguistic criteria.

%% file: parts/5_limitations.tex
\textbf{Robustness.} In the robustness experiments, the
\textit{Copeland} and \textit{Minimax} rules are less sensitive to performance score drops than $\sigma^{am}$ and $\sigma^{og}$. However, in certain circumstances \textit{Minimax} may display low resistance to such corruption due to its nature, which is analysed in \S\ref{subsec:missing_scores}. Other robustness evaluation settings can be considered, such as sensitivity to removing and adding new tasks~\citep{procaccia2007robustness,colombobest}, which are out of scope of this work.

\vspace{0.07em} \noindent \textbf{Ambiguity.} Almost all rules in our study allow ties or the recognition of systems as equivalent. This may result in non-resoluteness: the selection of multiple winning systems or the presence of many equivalencies in ranking. However, we empirically observe no or a few ties using the \textit{Dowdall}, \textit{Borda}, and \textit{Copeland} rules, whilst \textit{Minimax} and \textit{Plurality} treat a significant number of systems as equivalent due to their properties (\S\ref{subsection:reranking-benchmarks}). \textsc{Vote'n'Rank} does not currently support any additional tie-breaking rules to be applied in this case. The only exception here is the \textit{Threshold rule} that gives only one winner in almost all cases due to the built-in tie-breaking procedure.

\vspace{0.07em} \noindent  \textbf{Independence of irrelevant alternatives.} The violation of the IIA axiom in applications is a well-known fundamental aspect in the social choice theory, and the voting rules can violate the IIA with different probabilities~\cite{dougherty2020probability}. IIA violation may imply undesirable behavior: submitting a new system to the leaderboard affects the relative ranking of the other systems. However, we empirically show that \textit{Copeland} and \textit{Minimax} are less likely to violate IIA than \textit{Plurality} and \textit{Borda} rules (\S\ref{subsection:reranking-benchmarks}). The IIA assumption may be unrealistic in practice as it takes no account of perfect or near-perfect substitutes~\cite{suppes1965preference}.

\vspace{0.07em} \noindent  \textbf{Lack of ground truth.} Comparison of the aggregation procedures is hindered by the absence of the correct ranking, especially when votes are noisy and incomplete. There is no universal answer to the question of how the systems on the multi-task benchmarks should be preferred. However, we hope to contribute from a practical standpoint, offering an alternative approach to the mean aggregation procedure.

%% file: parts/appendix_statement.tex
Stereotypes and discrimination in LMs' pre-training data can lead to representation biases against race, religion, and social minorities. Our framework allows ranking systems to account for sensitive attributes~\citep{celis2018ranking}, e.g. gender and nationality, or to find the trade-off between multiple criteria, e.g. performance and fairness~\citep{baldini-etal-2022-fairness}. The rank aggregation rules have been widely adopted to information retrieval and recommendation systems~\citep{dwork2001rank,masthoff2011group}. We assume that translation of the social choice theory into the system evaluation problems may improve the user experience by selecting systems that best satisfy evaluative criteria and individual or group preferences in downstream applications.

%% file: parts/acknowledgement.tex
Mark Rofin, Mikhail Florinskiy, and Daniel Karabekyan were supported by the grant for research centres in the field of AI provided by the Analytical Centre for the Government of the Russian Federation (ACRF) in accordance with the agreement on the provision of subsidies (identifier of the agreement 000000D730321P5Q0002) and the agreement with the HSE University No. 70-2021-00139. We acknowledge the computational resources of HPC facilities at the HSE University. We would also like to thank colleagues from the IT University of Copenhagen and the anonymous reviewers for their comments on this paper.

%% file: parts/appendix_example_rules.tex
\subsection{Examples}
\label{appendix_a:examples}
\begin{table}[ht!]
    \centering
    \resizebox{1\columnwidth}{!}{
    \input{tables/toy_ranking_1}
    }
    \caption{A toy leaderboard for illustration purposes.}
    \label{tab:toy_ranking_1}
\end{table}

\begin{table}[ht!]
    \centering
    \resizebox{1\columnwidth}{!}{
    \input{tables/toy_ranking_2}
    }
    \caption{The leaderboard based on~\autoref{tab:toy_ranking_1} used for describing the \textit{Baldwin} rule.} \label{tab:toy_ranking_2}
\end{table}

This appendix provides illustrative examples on how our voting rules work. Here, suppose we have a toy leaderboard with five tasks and four systems as shown in~\autoref{tab:toy_ranking_1}. The systems are ranked within each task by their performance score. We now compute the rankings using each voting rule.

\paragraph{Scoring rules.}

\begin{itemize}
    \item \textit{Plurality rule} assigns the score of $2$ to $m_A$ and scores of $1$ to $m_B$, $m_C$, and $m_D$.
    \item According to the \textit{Borda rule}, the systems that take the first position get $3$ points for each task, $2$ points are awarded for the second position, etc. As a result, the systems receive the following \emph{Borda} scores: $m_A=6$, $m_B=9$, $m_C=8$, and $m_D=7$. The system $m_B$ has the highest score and is chosen as the best one.
    \item For the \textit{Dowdall} rule scoring vector, we get the following scores: $S_{m_A}=2.75$, $S_{m_B}=2.75$, $S_{m_C}=2.5$, and $S_{m_D}=1.75 + 2/3$. There is a tie between the systems $m_A$ and $m_B$, and both of them are considered the best models. 
\end{itemize}

\paragraph{Iterative scoring rules.}
\begin{itemize}
    \item For the \emph{Threshold} rule scoring vector (1,1,1,0), we get the following scores: $S_{m_A}=2$, $S_{m_B}=4$, $S_{m_C}=5$, and $S_{m_D}=4$. The system $m_C$ is the winner. If there is a tie, the scoring vector (1,1,0,0) is further applied for only tied systems.
    \item The \textit{Baldwin} rule: first, we calculate the \emph{Borda} scores as  mentioned above. Second, we eliminate the system $m_A$ since it has the lowest score (see~\autoref{tab:toy_ranking_2}). Next, we re-calculate the \emph{Borda} scores for a new scoring vector (2,1,0) and get the following results: $S_{m_B}=6$, $S_{m_C}=5$, and $S_{m_D}=4$. At this step, the system $m_D$ is eliminated. Finally, we re-calculate the results for the scoring vector (1,0). The results are $S_{m_B}=3$ and $S_{m_C}=2$, and the system $m_B$ is declared the winner.
\end{itemize}

\begin{figure}[t!]
    \centering
    \includegraphics[width=0.95\columnwidth]{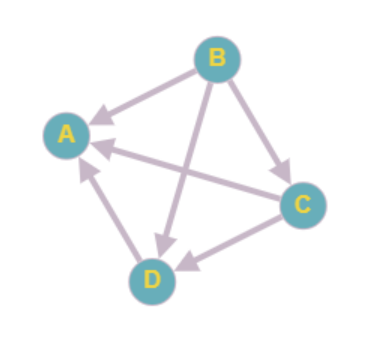}
    \caption{A toy graph example of the majority relation $\mu$ based on~\autoref{tab:toy_ranking_1}.} \label{fig:ex_graph}
\end{figure}

\paragraph{Majority-relation based rules.} The majority relation in this example is illustrated in~\autoref{fig:ex_graph}.
\begin{itemize}
    \item The system $m_B$ is the \textit{Condorcet} winner as it beats each of the alternatives. Note that since all majority-relation based rules (\textit{Copeland} and \textit{Minimax}) are \emph{Condorcet consistent}, they declare the system $m_B$ the winner as well. Let us illustrate it in more detail.
    \item The \textit{Copeland} rule scores are $u(m_A)=-3$, $u(m_B)=3$, $u(m_C)=1$, $u(m_D)=-1$. The system $m_B$ is the winner as it has the highest $u(x)$.
    \item The \textit{Minimax} rule scores are $rank(m_A)=-3$, $rank(m_B)=0$, $rank(m_C)=-3$, $rank(m_D)=-3$. Here, the system $m_B$ has the highest rank.
\end{itemize}

\subsection{Properties}
\label{appendix_a:properties}
We consider the following properties to describe our voting rules and summarise them in~\autoref{tab:properties}.

\begin{table}[t!]
    \centering
    \resizebox{1\columnwidth}{!}{
    \input{tables/properties}
    }
    \caption{Rules and their properties. $^*$The non-winning Pareto-dominated systems can be tied.}
    \label{tab:properties}
\end{table}

\begin{itemize}
    \item \textbf{Transitivity.} There are no cycles in the final ranking. An example of the cycle is a situation, where $m_A$ is better than $m_B$, $m_B$ is better than $m_C$, and $m_C$ is better than $m_A$. 
    \item \textbf{Unanimity (Pareto efficacy).} If the system $m_A$ is ranked higher than $m_B$ according to all criteria, then $m_A$ should be ranked higher.
    \item \textbf{Non-dictatorship (Anonymity).} There is no single criterion that defines the final ranking.
    \item \textbf{Independence of irrelevant alternatives (IIA).} For any two systems, the information about other systems should not influence their ranking.
    \item \textbf{Monotonicity.} If the system $m_A$ is the winner and it started to rank higher according to one of the criteria, then it should still be the winner.
    \item \textbf{Majority criterion.} If the system $m_A$ is considered the best by more than 50\% criteria, then it should be the winner.
    \item \textbf{Condorcet winner criterion}. This criterion is a stronger version of the Majority criterion. If the system $m_A$ is the Condorcet winner (CW), it should be the winner according to the rule.
    \item \textbf{Condorcet loser criterion.} If the system $m_A$ is the Condorcet loser ($a \mu M^L$ for any $a \in A$), it should never be the winner according to the rule.
\end{itemize}

Recall that the \emph{Condorcet} rule by definition complies with the Condorcet winner and loser criteria. The other properties can not be checked in application to benchmarking since the rule is defined on a restricted domain: it does not provide the results on any possible combination of rankings.

There is no single best voting rule since none of them satisfies properties of the Arrow's impossibility theorem~\cite{arrow2012social,geanakoplos2005three}: transitivity, unanimity, non-dictatorship, and independence of irrelevant alternatives (IIA). 

%% file: tables/toy_ranking_1.tex
\begin{tabular}{llllll}
\toprule
Rank & Task 1 & Task 2 & Task 3 & Task 4 & Task 5 \\ \midrule
1    & $m_A$      & $m_A$      & $m_B$      & $m_C$      & $m_D$      \\
2    & $m_B$      & $m_C$      & $m_D$      & $m_B$      & $m_B$      \\
3    & $m_C$      & $m_D$      & $m_C$      & $m_D$      & $m_C$      \\
4    & $m_D$      & $m_B$      & $m_A$      & $m_A$      & $m_A$      \\
\bottomrule
\end{tabular}

%% file: tables/toy_ranking_2.tex
\begin{tabular}{llllll}
\toprule
Rank & Task 1 & Task 2 & Task 3 & Task 4 & Task 5 \\
\midrule
1    & $m_B$      & $m_C$      & $m_B$      & $m_C$      & $m_D$      \\
2    & $m_C$      & $m_D$      & $m_D$      & $m_B$      & $m_B$      \\
3    & $m_D$      & $m_B$      & $m_C$      & $m_D$      & $m_C$ \\ \bottomrule
\end{tabular}

%% file: tables/properties.tex
\setlength{\tabcolsep}{2pt}
\begin{tabular}{lccccccc}
\toprule
& \rotatebox{65}{\textbf{Plurality}} & \rotatebox{65}{\textbf{Borda}} & \rotatebox{65}{\textbf{Dowdall}} & \rotatebox{65}{\textbf{Threshold}} & \rotatebox{65}{\textbf{Baldwin}} & \rotatebox{65}{\textbf{Copeland}} & \rotatebox{65}{\textbf{Minmax}} \\
\midrule
{Transitivity} & \checkmark & \checkmark & \checkmark & \checkmark & \checkmark & \checkmark & \checkmark \\
{Anonymity} & \checkmark & \checkmark & \checkmark & \checkmark & \checkmark & \checkmark & \checkmark \\
{Unanimity} & \checkmark$^*$ & \checkmark & \checkmark & \checkmark & \checkmark & \checkmark & \checkmark \\
{IIA} & \xmark & \xmark & \xmark & \xmark & \xmark & \xmark & \xmark \\
{Monotonicity} & \checkmark & \checkmark & \checkmark & \checkmark & \xmark & \checkmark & \checkmark \\
{Majority} & \checkmark & \xmark & \checkmark & \xmark & \checkmark & \checkmark & \checkmark \\
{CW} & \xmark & \xmark & \xmark & \xmark & \checkmark & \checkmark & \checkmark \\
{Condorcet loser} &  \xmark & \checkmark & \xmark & \xmark & \checkmark & \checkmark & \xmark \\
\midrule



Sum & 5 & 5 & 5 & 4 & 6 & 7 & 6 \\
\bottomrule
\end{tabular}

%% file: parts/appendix_case_studies.tex
We do not report the agreement rate, the Kendall Tau ($\tau$) correlation, and the IIA results for \textsc{VALUE} since we are given only up to $7$ evaluated alternatives: \textsc{craig.starr}~\citep{shin2021winning}, \textsc{DuKG}~\citep{li2021clip}, \textsc{Human}, and four \textsc{HERO}-based configurations~\citep{li-etal-2020-hero}. The \textsc{HERO}-based baselines are trained in the following settings: single-task training (ST), multi-task training (MT) by tasks or domains, all-task training (AT) and AT first then ST (AT ->
ST). We refer to the configurations as follows:
\begin{itemize}[noitemsep]
    \item HERO$_1$: AT->ST, PT+FT;
    \item HERO$_2$: AT->ST, FT-only;
    \item HERO$_3$: ST, PT+FT;
    \item HERO$_4$: ST, FT-only.
\end{itemize}

\vspace{0.7em}\noindent\textbf{The VALUE results.}  \autoref{tab:value_cs1} and \autoref{tab:value_cs1_non} show the \textsc{VALUE} re-ranking results over missing/non-missing scores. \xmark~means that the given aggregation method does not operate over missing values. In the first case, we observe that the \textit{Copeland} and \textit{Minimax} rules generally agree on the final outcomes except for the fifth and sixth positions. The rules select \textsc{Human} as the winner. At the same time   \textsc{DuKG} and \textsc{HERO$_1$} have the same \textit{Minimax} values, and the \textit{Minimax} values of the least-$3$ systems are also equal. In the second case, we omit \textsc{Human} due to missing scores on $6$ out of $11$ tasks for comparable interpretation. However, there are $3$ tied alternatives in the \textit{Minimax} and \textit{Dowdall} outcomes. Interestingly, all methods are consistent in conclusions on the top-$3$ systems, with the \textit{Minimax} treating \textsc{DuKG} and \textsc{HERO$_1$} as equal alternatives. The $\sigma^{og}$, \textit{Minimax}, \textit{Plurality}, \textit{Dowdall}, and \textit{Borda} rules make equal decisions on the final outcomes. 

\input{tables/superglue_cs1}
\input{tables/value_cs1}

\input{tables/value_cs1_non}

%% file: tables/superglue_cs1.tex
\begin{table*}[!ht]
\centering
\resizebox{1\linewidth}{!}{
\scriptsize
\begin{tabular}{ccccccccc}
\toprule
\textbf{Rank} & $\sigma^{am}$ & $\sigma^{gm}$ & $\sigma^{og}$ & \textbf{Copeland} & \textbf{Minimax} & \textbf{Plurality} & \textbf{Dowdall} & \textbf{Borda}\\  \midrule

1&  \ernie$^{90.62}$ &  \ernie$_{\updownarrow0}^{90.04}$ & \ernie$_{\updownarrow0}^{0.066}$&\humanpersoni$_{\uparrow3}^{20.00}$ &\ernie$_{\updownarrow0}^{0}$ &\humanpersoni$_{\uparrow3}^{4.00}$ &\humanpersoni$_{\uparrow3}^{4.98}$ &  \ernie$_{\updownarrow0}^{155.00}$ \\
2&  \johnson$^{90.39}$ &  \johnson$_{\updownarrow0}^{89.84}$ & \johnson$_{\updownarrow0}^{0.068}$&  \ernie$_{\downarrow1}^{19.00}$ &\humanpersoni$_{\uparrow2}^{0}$ &  \ernie$_{\downarrow1}^{2.50}$ &  \ernie$_{\downarrow1}^{4.25}$ &  \johnson$_{\updownarrow0}^{154.50}$ \\
3&  \ovejita$^{90.29}$ &  \ovejita$_{\updownarrow0}^{89.75}$ & \ovejita$_{\updownarrow0}^{0.068}$& \johnson$_{\downarrow1}^{18.00}$ &  \johnson$_{\downarrow1}^{-4.50}$ &  \johnson$_{\downarrow1}^{1.00}$ &  \johnson$_{\downarrow1}^{3.62}$ &  \ovejita$_{\updownarrow0}^{153.00}$ \\
4&\humanpersoni$^{89.79}$ &\humanpersoni$_{\updownarrow0}^{88.80}$ & \slimey$_{\uparrow1}^{0.073}$& \ovejita$_{\downarrow1}^{15.00}$ &  \ovejita$_{\downarrow1}^{-5.00}$ &  \ovejita$_{\downarrow1}^{0.50}$ &  \ovejita$_{\downarrow1}^{3.29}$ &\humanpersoni$_{\updownarrow0}^{145.50}$ \\
5& \slimey$^{89.25}$ & \slimey$_{\updownarrow0}^{88.75}$ & \humanpersoni$_{\downarrow1}^{0.074}$  & \slimey$_{\updownarrow0}^{13.00}$ & \slimey$_{\updownarrow0}^{-7.50}$ &   \sam$_{\uparrow9}^{0.00}$ & \slimey$_{\updownarrow0}^{2.11}$ & \slimey$_{\updownarrow0}^{141.50}$ \\
6& \betty$^{86.65}$ & \betty$_{\updownarrow0}^{85.93}$ & \frazzle$_{\uparrow1}^{0.089}$ & \betty$_{\updownarrow0}^{11.00}$ &   \sam$_{\uparrow8}^{-8.00}$ &\bert$_{\uparrow12}^{0.00}$ & \betty$_{\updownarrow0}^{1.16}$ & \betty$_{\updownarrow0}^{116.50}$ \\
7& \frazzle$^{86.09}$ & \frazzle$_{\updownarrow0}^{85.38}$ & \bday$_{\uparrow5}^{0.10}$ & \prairie$_{\uparrow1}^{9.00}$ &\bert$_{\uparrow11}^{-8.00}$ & \barkley$_{\uparrow12}^{0.00}$ &\prairie$_{\uparrow1}^{1.06}$ &  \prairie$_{\uparrow1}^{108.00}$ \\
\bottomrule
\end{tabular}
}
\caption{Results of re-ranking the \textsc{SGLUE} benchmark. The model rank changes are depicted with arrows, whilst the superscripts denote scores assigned by the voting method. \underline{Notations}: \humanpersoni=\textsc{Human}; \ernie=\textsc{ERNIE};  \ovejita=\textsc{DeBERTa/TuringNLRv4};
\betty=\textsc{NEZHA-Plus};
\johnson=\textsc{T5+UDG};
\bert=\textsc{BERT++};
\frazzle=\textsc{PAI Albert};
\prairie=\textsc{RoBERTa-iCETS};
\barkley=\textsc{GPT-3 few-shot};
\sam=\textsc{iPET (ALBERT) few-shot};
\slimey=\textsc{T5};\bday=\textsc{AILabs Team}.}
\label{superglue_cs1}
\end{table*}

%% file: tables/value_cs1.tex
\begin{table*}[h!]
\centering
\resizebox{1\linewidth}{!}{
\scriptsize
\begin{tabular}{ccccccccc}
\toprule
\textbf{Rank} & $\sigma^{am}$ & $\sigma^{gm}$ & $\sigma^{og}$ &\textbf{Copeland} & \textbf{Minimax} & \textbf{Plurality} & \textbf{Dowdall} & \textbf{Borda}\\ \midrule
1  &  \xmark &   \xmark & \xmark&  \humanpersonj$_{\updownarrow0}^{6.00}$ & \humanpersonj$_{\updownarrow0}^{0}$ & \xmark & \xmark & \xmark  \\
2  &  \xmark &   \xmark &\xmark&\sherlockvalue$_{\updownarrow0}^{4.00}$ & \sherlockvalue$_{\updownarrow0}^{-6.00}$ & \xmark & \xmark & \xmark \\
3  & \xmark &   \xmark  & \xmark&  \barkley$_{\updownarrow0}^{2.00}$ &\barkley$_{\updownarrow0}^{-10.00}$ & \xmark & \xmark & \xmark \\
4  &\xmark &   \xmark &\xmark& \frazzle$_{\updownarrow0}^{0.00}$ &  \frazzle$_{\updownarrow0}^{-10.00}$ & \xmark & \xmark & \xmark \\
5  &\xmark &   \xmark &\xmark&\sam$_{\updownarrow0}^{-2.00}$ &\johnson$_{\updownarrow0}^{-11.00}$ & \xmark & \xmark & \xmark \\
6  & \xmark &   \xmark & \xmark& \johnson$_{\updownarrow0}^{-4.00}$ &\sam$_{\updownarrow0}^{-11.00}$ & \xmark & \xmark & \xmark \\
7  & \xmark &   \xmark & \xmark& \mumford$_{\updownarrow0}^{-6.00}$ &\mumford$_{\updownarrow0}^{-11.00}$ & \xmark & \xmark & \xmark \\
\bottomrule
\end{tabular}
}
\caption{Results of re-ranking the \textsc{VALUE} benchmark over missing scores. Changes in the system ranks are depicted with arrows, whilst the superscripts denote scores assigned by the aggregation procedure. \underline{Notations}: \humanpersonj=\textsc{Human}; \sherlockvalue=\textsc{craig.starr}; \barkley=\textsc{DuKG}; \frazzle=\textsc{HERO$_1$}; \sam=\textsc{HERO$_2$}; \johnson=\textsc{HERO$_3$}; \mumford=\textsc{HERO$_4$}.
}
\label{tab:value_cs1}
\end{table*}

%% file: tables/value_cs1_non.tex
\begin{table*}[h!]
\centering
\resizebox{1\linewidth}{!}{
\scriptsize
\begin{tabular}{ccccccccc}
\toprule
\textbf{Rank} & $\sigma^{am}$ & $\sigma^{gm}$ & $\sigma^{og}$ & \textbf{Copeland} & \textbf{Minimax} & \textbf{Plurality} & \textbf{Dowdall} & \textbf{Borda}\\ \midrule

1                &             \sherlockvalue$^{62.87}$ &             \sherlockvalue$_{\updownarrow0}^{49.96}$ &  \sherlockvalue$_{\updownarrow0}^{0.365}$&            \sherlockvalue$_{\updownarrow0}^{5.00}$ &            \sherlockvalue$_{\updownarrow0}^{0}$ &           \sherlockvalue$_{\updownarrow0}^{9.00}$ &          \sherlockvalue$_{\updownarrow0}^{10.00}$ &           \sherlockvalue$_{\updownarrow0}^{53.00}$ \\
2                &                    \barkley$^{60.00}$ &                    \barkley$_{\updownarrow0}^{46.30}$ &  \barkley$_{\updownarrow0}^{0.381}$  &                  \barkley$_{\updownarrow0}^{3.00}$ &                  \barkley$_{\updownarrow0}^{-10.00}$ &                  \barkley$_{\updownarrow0}^{1.00}$ &                  \barkley$_{\updownarrow0}^{5.17}$ &                  \barkley$_{\updownarrow0}^{39.00}$ \\
3                &    \frazzle$^{57.58}$ &    \frazzle$_{\updownarrow0}^{44.12}$ & 
\frazzle$_{\updownarrow0}^{0.399}$ &
\frazzle$_{\updownarrow0}^{1.00}$ &  \frazzle$_{\updownarrow0}^{-10.00}$ &  \frazzle$_{\updownarrow0}^{1.00}$ &  \frazzle$_{\updownarrow0}^{4.08}$ &  \frazzle$_{\updownarrow0}^{30.00}$ \\
4                &        \sam$^{56.96}$ &        \sam$_{\updownarrow0}^{43.22}$ &
\johnson$_{\uparrow1}^{0.403}$ &
\sam$_{\updownarrow0}^{-1.00}$ &    \johnson$_{\uparrow1}^{-11.00}$ &    \johnson$_{\uparrow1}^{0.00}$ &    \johnson$_{\uparrow1}^{2.87}$ &    \johnson$_{\uparrow1}^{20.00}$ \\
5                &  \johnson$^{56.07}$ &  \johnson$_{\updownarrow0}^{42.81}$ & 
\sam$_{\downarrow1}^{0.404}$ & 
\johnson$_{\updownarrow0}^{-3.00}$ &        \sam$_{\downarrow1}^{-11.00}$ &        \sam$_{\downarrow1}^{0.00}$ &        \sam$_{\downarrow1}^{2.82}$ &        \sam$_{\downarrow1}^{18.00}$ \\
6                &      \mumford$^{52.59}$ &      \mumford$_{\updownarrow0}^{37.56}$ &  
\mumford$_{\updownarrow0}^{0.438}$ &\mumford$_{\updownarrow0}^{-5.00}$ &    \mumford$_{\updownarrow0}^{-11.00}$ &    \mumford$_{\updownarrow0}^{0.00}$ &    \mumford$_{\updownarrow0}^{2.02}$ &     \mumford$_{\updownarrow0}^{5.00}$ \\

\bottomrule
\end{tabular}
}
\caption{Results of re-ranking the \textsc{VALUE} benchmark over non-missing scores. The \textsc{Human} results are discarded due to missing scores. Changes in the system ranks are depicted with arrows, whilst the superscripts denote scores assigned by the aggregation procedure. \underline{Notations}: \sherlockvalue=\textsc{craig.starr}; \barkley=\textsc{DuKG}; \frazzle=\textsc{HERO$_1$}; \sam=\textsc{HERO$_2$}; \johnson=\textsc{HERO$_3$}; \mumford=\textsc{HERO$_4$}.
}
\label{tab:value_cs1_non}
\end{table*}

%% file: main.bbl
\begin{thebibliography}{74}
\expandafter\ifx\csname natexlab\endcsname\relax\def\natexlab#1{#1}\fi

\bibitem[{Agarwal et~al.(2021)Agarwal, Schwarzer, Castro, Courville, and
  Bellemare}]{agarwal2021deep}
Rishabh Agarwal, Max Schwarzer, Pablo~Samuel Castro, Aaron~C Courville, and
  Marc Bellemare. 2021.
\newblock {Deep Reinforcement Learning at the Edge of the Statistical
  Precipice}.
\newblock \emph{Advances in Neural Information Processing Systems}, 34.

\bibitem[{Aizerman and Aleskerov(1995)}]{auizerman1995theory}
Mark Aizerman and Fuad Aleskerov. 1995.
\newblock \emph{{Theory of Choice}}, volume~38.
\newblock North Holland.

\bibitem[{Aleskerov et~al.(2010)Aleskerov, Chistyakov, and
  Kalyagin}]{aleskerov2010threshold}
Fuad Aleskerov, Vyacheslav~V Chistyakov, and Valery Kalyagin. 2010.
\newblock The threshold aggregation.
\newblock \emph{Economics Letters}, 107(2):261--262.

\bibitem[{Arrow(2012)}]{arrow2012social}
Kenneth~J Arrow. 2012.
\newblock Social choice and individual values.
\newblock In \emph{Social Choice and Individual Values}. Yale university press.

\bibitem[{Baldini et~al.(2022)Baldini, Wei, Natesan~Ramamurthy, Singh, and
  Yurochkin}]{baldini-etal-2022-fairness}
Ioana Baldini, Dennis Wei, Karthikeyan Natesan~Ramamurthy, Moninder Singh, and
  Mikhail Yurochkin. 2022.
\newblock \href {https://doi.org/10.18653/v1/2022.findings-acl.176} {Your
  fairness may vary: Pretrained language model fairness in toxic text
  classification}.
\newblock In \emph{Findings of the Association for Computational Linguistics:
  ACL 2022}, pages 2245--2262, Dublin, Ireland. Association for Computational
  Linguistics.

\bibitem[{Bartholdi et~al.(1989)Bartholdi, Tovey, and
  Trick}]{bartholdi1989voting}
John Bartholdi, Craig~A Tovey, and Michael~A Trick. 1989.
\newblock Voting schemes for which it can be difficult to tell who won the
  election.
\newblock \emph{Social Choice and welfare}, 6(2):157--165.

\bibitem[{Belz et~al.(2021)Belz, Agarwal, Shimorina, and
  Reiter}]{belz-etal-2021-systematic}
Anya Belz, Shubham Agarwal, Anastasia Shimorina, and Ehud Reiter. 2021.
\newblock \href {https://doi.org/10.18653/v1/2021.eacl-main.29} {A systematic
  review of reproducibility research in natural language processing}.
\newblock In \emph{Proceedings of the 16th Conference of the European Chapter
  of the Association for Computational Linguistics: Main Volume}, pages
  381--393, Online. Association for Computational Linguistics.

\bibitem[{Benavoli et~al.(2016)Benavoli, Corani, and
  Mangili}]{JMLR:v17:benavoli16a}
Alessio Benavoli, Giorgio Corani, and Francesca Mangili. 2016.
\newblock \href {http://jmlr.org/papers/v17/benavoli16a.html} {{Should We
  Really Use Post-Hoc Tests Based on Mean-Ranks?}}
\newblock \emph{Journal of Machine Learning Research}, 17(5):1--10.

\bibitem[{Bender et~al.(2021)Bender, Gebru, McMillan-Major, and
  Shmitchell}]{bender2021dangers}
Emily~M Bender, Timnit Gebru, Angelina McMillan-Major, and Shmargaret
  Shmitchell. 2021.
\newblock {On the Dangers of Stochastic Parrots: Can Language Models Be Too
  Big?}
\newblock In \emph{Proceedings of the 2021 ACM Conference on Fairness,
  Accountability, and Transparency}, pages 610--623.

\bibitem[{Black et~al.(1958)}]{black1958theory}
Duncan Black et~al. 1958.
\newblock The theory of committees and elections.

\bibitem[{Bowman and Dahl(2021)}]{bowman-dahl-2021-will}
Samuel~R. Bowman and George Dahl. 2021.
\newblock \href {https://doi.org/10.18653/v1/2021.naacl-main.385} {What will it
  take to fix benchmarking in natural language understanding?}
\newblock In \emph{Proceedings of the 2021 Conference of the North American
  Chapter of the Association for Computational Linguistics: Human Language
  Technologies}, pages 4843--4855, Online. Association for Computational
  Linguistics.

\bibitem[{Brandt and Seedig(2016)}]{brandt2016discriminative}
Felix Brandt and Hans~Georg Seedig. 2016.
\newblock {On the Discriminative Power of Tournament Solutions}.
\newblock In \emph{Operations Research Proceedings 2014}, pages 53--58.
  Springer.

\bibitem[{Celis et~al.(2018)Celis, Straszak, and Vishnoi}]{celis2018ranking}
L~Elisa Celis, Damian Straszak, and Nisheeth~K Vishnoi. 2018.
\newblock {Ranking with Fairness Constraints}.
\newblock In \emph{45th International Colloquium on Automata, Languages, and
  Programming (ICALP 2018)}. Schloss Dagstuhl-Leibniz-Zentrum fuer Informatik.

\bibitem[{Choudhury and Deshpande(2021)}]{choudhury2021linguistically}
Monojit Choudhury and Amit Deshpande. 2021.
\newblock {How Linguistically Fair are Multilingual Pre-trained Language
  Models}.
\newblock In \emph{Proceedings of the AAAI Conference on Artificial
  Intelligence}, volume~35, pages 12710--12718.

\bibitem[{Clark et~al.(2019)Clark, Lee, Chang, Kwiatkowski, Collins, and
  Toutanova}]{clark-etal-2019-boolq}
Christopher Clark, Kenton Lee, Ming-Wei Chang, Tom Kwiatkowski, Michael
  Collins, and Kristina Toutanova. 2019.
\newblock \href {https://doi.org/10.18653/v1/N19-1300} {{B}ool{Q}: Exploring
  the surprising difficulty of natural yes/no questions}.
\newblock In \emph{Proceedings of the 2019 Conference of the North {A}merican
  Chapter of the Association for Computational Linguistics: Human Language
  Technologies, Volume 1 (Long and Short Papers)}, pages 2924--2936,
  Minneapolis, Minnesota. Association for Computational Linguistics.

\bibitem[{Colombo et~al.()Colombo, Noiry, Irurozki, and
  CLEMENCON}]{colombobest}
Pierre Colombo, Nathan Noiry, Ekhine Irurozki, and Stephan CLEMENCON.
\newblock {What are the Best Systems? New Perspectives on NLP Benchmarking}.
\newblock In \emph{Advances in Neural Information Processing Systems}.

\bibitem[{Colombo et~al.(2022)Colombo, Clavel, and
  Piantanida}]{colombo2022infolm}
Pierre Jean~A Colombo, Chlo{\'e} Clavel, and Pablo Piantanida. 2022.
\newblock {InfoLM: A New Metric to Evaluate Summarization \& Data2Text
  Generation}.
\newblock In \emph{Proceedings of the AAAI Conference on Artificial
  Intelligence}, volume~36, pages 10554--10562.

\bibitem[{De~Almeida et~al.(2019)De~Almeida, Morais, and Nurmi}]{de2019systems}
Adiel~Teixeira De~Almeida, Danielle~Costa Morais, and Hannu Nurmi. 2019.
\newblock \emph{Systems, procedures and voting rules in context: A primer for
  voting rule selection}, volume~9.
\newblock Springer.

\bibitem[{Dehghani et~al.(2021)Dehghani, Tay, Gritsenko, Zhao, Houlsby, Diaz,
  Metzler, and Vinyals}]{dehghani2021benchmark}
Mostafa Dehghani, Yi~Tay, Alexey~A Gritsenko, Zhe Zhao, Neil Houlsby, Fernando
  Diaz, Donald Metzler, and Oriol Vinyals. 2021.
\newblock {The Benchmark Lottery}.
\newblock \emph{arXiv preprint arXiv:2107.07002}.

\bibitem[{Dem{\v{s}}ar(2006)}]{demvsar2006statistical}
Janez Dem{\v{s}}ar. 2006.
\newblock {Statistical Comparisons of Classifiers over Multiple Data Sets}.
\newblock \emph{The Journal of Machine learning research}, 7:1--30.

\bibitem[{Devlin et~al.(2019)Devlin, Chang, Lee, and
  Toutanova}]{devlin-etal-2019-bert}
Jacob Devlin, Ming-Wei Chang, Kenton Lee, and Kristina Toutanova. 2019.
\newblock \href {https://doi.org/10.18653/v1/N19-1423} {{BERT}: Pre-training of
  deep bidirectional transformers for language understanding}.
\newblock In \emph{Proceedings of the 2019 Conference of the North {A}merican
  Chapter of the Association for Computational Linguistics: Human Language
  Technologies, Volume 1 (Long and Short Papers)}, pages 4171--4186,
  Minneapolis, Minnesota. Association for Computational Linguistics.

\bibitem[{Dougherty and Heckelman(2020)}]{dougherty2020probability}
Keith~L Dougherty and Jac~C Heckelman. 2020.
\newblock The probability of violating arrow’s conditions.
\newblock \emph{European Journal of Political Economy}, 65:101936.

\bibitem[{Dwork et~al.(2001)Dwork, Kumar, Naor, and Sivakumar}]{dwork2001rank}
Cynthia Dwork, Ravi Kumar, Moni Naor, and Dandapani Sivakumar. 2001.
\newblock {Rank Aggregation Methods for the Web}.
\newblock In \emph{Proceedings of the 10th international conference on World
  Wide Web}, pages 613--622.

\bibitem[{Edelman(2015)}]{edelman2015myth}
Paul~H Edelman. 2015.
\newblock The myth of the condorcet winner.
\newblock \emph{Supreme Court Economic Review}, 22(1):207--219.

\bibitem[{Elangovan et~al.(2021)Elangovan, He, and
  Verspoor}]{elangovan-etal-2021-memorization}
Aparna Elangovan, Jiayuan He, and Karin Verspoor. 2021.
\newblock \href {https://doi.org/10.18653/v1/2021.eacl-main.113} {Memorization
  vs. generalization : Quantifying data leakage in {NLP} performance
  evaluation}.
\newblock In \emph{Proceedings of the 16th Conference of the European Chapter
  of the Association for Computational Linguistics: Main Volume}, pages
  1325--1335, Online. Association for Computational Linguistics.

\bibitem[{Ethayarajh and Jurafsky(2020)}]{ethayarajh-jurafsky-2020-utility}
Kawin Ethayarajh and Dan Jurafsky. 2020.
\newblock \href {https://doi.org/10.18653/v1/2020.emnlp-main.393} {Utility is
  in the eye of the user: A critique of {NLP} leaderboards}.
\newblock In \emph{Proceedings of the 2020 Conference on Empirical Methods in
  Natural Language Processing (EMNLP)}, pages 4846--4853, Online. Association
  for Computational Linguistics.

\bibitem[{Felsenthal and Machover(2012)}]{felsenthal2012electoral}
Dan~S Felsenthal and Mosh{\'e} Machover. 2012.
\newblock \emph{{Electoral Systems: Paradoxes, Assumptions, and Procedures}}.
\newblock Springer Science \& Business Media.

\bibitem[{Geanakoplos(2005)}]{geanakoplos2005three}
John Geanakoplos. 2005.
\newblock Three brief proofs of arrow’s impossibility theorem.
\newblock \emph{Economic Theory}, 26(1):211--215.

\bibitem[{He et~al.(2020)He, Liu, Gao, and Chen}]{he2020deberta}
Pengcheng He, Xiaodong Liu, Jianfeng Gao, and Weizhu Chen. 2020.
\newblock {DeBERTa: decoding-enhanced BERT with disentangled attention}.
\newblock In \emph{International Conference on Learning Representations}.

\bibitem[{Henderson et~al.(2020)Henderson, Hu, Romoff, Brunskill, Jurafsky, and
  Pineau}]{henderson2020towards}
Peter Henderson, Jieru Hu, Joshua Romoff, Emma Brunskill, Dan Jurafsky, and
  Joelle Pineau. 2020.
\newblock {Towards the Systematic Reporting of the Energy and Carbon Footprints
  of Machine Learning}.
\newblock \emph{Journal of Machine Learning Research}, 21(248):1--43.

\bibitem[{Khashabi et~al.(2018)Khashabi, Chaturvedi, Roth, Upadhyay, and
  Roth}]{khashabi-etal-2018-looking}
Daniel Khashabi, Snigdha Chaturvedi, Michael Roth, Shyam Upadhyay, and Dan
  Roth. 2018.
\newblock \href {https://doi.org/10.18653/v1/N18-1023} {Looking beyond the
  surface: A challenge set for reading comprehension over multiple sentences}.
\newblock In \emph{Proceedings of the 2018 Conference of the North {A}merican
  Chapter of the Association for Computational Linguistics: Human Language
  Technologies, Volume 1 (Long Papers)}, pages 252--262, New Orleans,
  Louisiana. Association for Computational Linguistics.

\bibitem[{Kiela et~al.(2021)Kiela, Bartolo, Nie, Kaushik, Geiger, Wu, Vidgen,
  Prasad, Singh, Ringshia, Ma, Thrush, Riedel, Waseem, Stenetorp, Jia, Bansal,
  Potts, and Williams}]{kiela-etal-2021-dynabench}
Douwe Kiela, Max Bartolo, Yixin Nie, Divyansh Kaushik, Atticus Geiger,
  Zhengxuan Wu, Bertie Vidgen, Grusha Prasad, Amanpreet Singh, Pratik Ringshia,
  Zhiyi Ma, Tristan Thrush, Sebastian Riedel, Zeerak Waseem, Pontus Stenetorp,
  Robin Jia, Mohit Bansal, Christopher Potts, and Adina Williams. 2021.
\newblock \href {https://doi.org/10.18653/v1/2021.naacl-main.324} {Dynabench:
  Rethinking benchmarking in {NLP}}.
\newblock In \emph{Proceedings of the 2021 Conference of the North American
  Chapter of the Association for Computational Linguistics: Human Language
  Technologies}, pages 4110--4124, Online. Association for Computational
  Linguistics.

\bibitem[{Lan et~al.(2019)Lan, Chen, Goodman, Gimpel, Sharma, and
  Soricut}]{lan2019albert}
Zhenzhong Lan, Mingda Chen, Sebastian Goodman, Kevin Gimpel, Piyush Sharma, and
  Radu Soricut. 2019.
\newblock {ALBERT: A Lite BERT for self-supervised learning of language
  representations}.
\newblock In \emph{International Conference on Learning Representations}.

\bibitem[{Levin and Nalebuff(1995)}]{Levin-Nalebuff}
Jonathan Levin and Barry Nalebuff. 1995.
\newblock \href {http://www.jstor.org/stable/2138351} {An introduction to
  vote-counting schemes}.
\newblock \emph{The Journal of Economic Perspectives}, 9(1):3--26.

\bibitem[{Li et~al.(2021)Li, He, and Feng}]{li2021clip}
Guohao Li, Feng He, and Zhifan Feng. 2021.
\newblock {A CLIP-Enhanced Method for Video-Language Understanding}.
\newblock \emph{arXiv preprint arXiv:2110.07137}.

\bibitem[{Li et~al.(2020)Li, Chen, Cheng, Gan, Yu, and Liu}]{li-etal-2020-hero}
Linjie Li, Yen-Chun Chen, Yu~Cheng, Zhe Gan, Licheng Yu, and Jingjing Liu.
  2020.
\newblock \href {https://doi.org/10.18653/v1/2020.emnlp-main.161} {{HERO}:
  Hierarchical encoder for {V}ideo+{L}anguage omni-representation
  pre-training}.
\newblock In \emph{Proceedings of the 2020 Conference on Empirical Methods in
  Natural Language Processing (EMNLP)}, pages 2046--2065, Online. Association
  for Computational Linguistics.

\bibitem[{Li et~al.()Li, Lei, Gan, Yu, Chen, Pillai, Cheng, Zhou, Wang, Wang
  et~al.}]{li1value}
Linjie Li, Jie Lei, Zhe Gan, Licheng Yu, Yen-Chun Chen, Rohit Pillai, Yu~Cheng,
  Luowei Zhou, Xin~Eric Wang, William~Yang Wang, et~al.
\newblock {VALUE: A Multi-Task Benchmark for Video-and-Language Understanding
  Evaluation}.
\newblock In \emph{Thirty-fifth Conference on Neural Information Processing
  Systems Datasets and Benchmarks Track (Round 1)}.

\bibitem[{Liang et~al.(2020)Liang, Duan, Gong, Wu, Guo, Qi, Gong, Shou, Jiang,
  Cao, Fan, Zhang, Agrawal, Cui, Wei, Bharti, Qiao, Chen, Wu, Liu, Yang,
  Campos, Majumder, and Zhou}]{liang-etal-2020-xglue}
Yaobo Liang, Nan Duan, Yeyun Gong, Ning Wu, Fenfei Guo, Weizhen Qi, Ming Gong,
  Linjun Shou, Daxin Jiang, Guihong Cao, Xiaodong Fan, Ruofei Zhang, Rahul
  Agrawal, Edward Cui, Sining Wei, Taroon Bharti, Ying Qiao, Jiun-Hung Chen,
  Winnie Wu, Shuguang Liu, Fan Yang, Daniel Campos, Rangan Majumder, and Ming
  Zhou. 2020.
\newblock \href {https://doi.org/10.18653/v1/2020.emnlp-main.484} {{XGLUE}: A
  new benchmark datasetfor cross-lingual pre-training, understanding and
  generation}.
\newblock In \emph{Proceedings of the 2020 Conference on Empirical Methods in
  Natural Language Processing (EMNLP)}, pages 6008--6018, Online. Association
  for Computational Linguistics.

\bibitem[{Liu et~al.(2019)Liu, Ott, Goyal, Du, Joshi, Chen, Levy, Lewis,
  Zettlemoyer, and Stoyanov}]{liu2019roberta}
Yinhan Liu, Myle Ott, Naman Goyal, Jingfei Du, Mandar Joshi, Danqi Chen, Omer
  Levy, Mike Lewis, Luke Zettlemoyer, and Veselin Stoyanov. 2019.
\newblock {RoBERTa: A Robustly Optimized BERT Pretraining Approach}.
\newblock \emph{arXiv preprint arXiv:1907.11692}.

\bibitem[{Ma et~al.(2021)Ma, Ethayarajh, Thrush, Jain, Wu, Jia, Potts,
  Williams, and Kiela}]{dynaboard}
Zhiyi Ma, Kawin Ethayarajh, Tristan Thrush, Somya Jain, Ledell Wu, Robin Jia,
  Christopher Potts, Adina Williams, and Douwe Kiela. 2021.
\newblock \href
  {https://proceedings.neurips.cc/paper/2021/file/55b1927fdafef39c48e5b73b5d61ea60-Paper.pdf}
  {{Dynaboard: An Evaluation-As-A-Service Platform for Holistic Next-Generation
  Benchmarking}}.
\newblock In \emph{Advances in Neural Information Processing Systems},
  volume~34, pages 10351--10367. Curran Associates, Inc.

\bibitem[{Masthoff(2011)}]{masthoff2011group}
Judith Masthoff. 2011.
\newblock {Group Recommender Systems: Combining Individual Models}.
\newblock In \emph{Recommender systems handbook}, pages 677--702. Springer.

\bibitem[{Min et~al.(2021)Min, Boyd-Graber, Alberti, Chen, Choi, Collins, Guu,
  Hajishirzi, Lee, Palomaki, Raffel, Roberts, Kwiatkowski, Lewis, Wu,
  K\"uttler, Liu, Minervini, Stenetorp, Riedel, Yang, Seo, Izacard, Petroni,
  Hosseini, Cao, Grave, Yamada, Shimaoka, Suzuki, Miyawaki, Sato, Takahashi,
  Suzuki, Fajcik, Docekal, Ondrej, Smrz, Cheng, Shen, Liu, He, Chen, Gao, Oguz,
  Chen, Karpukhin, Peshterliev, Okhonko, Schlichtkrull, Gupta, Mehdad, and
  Yih}]{pmlr-v133-min21a}
Sewon Min, Jordan Boyd-Graber, Chris Alberti, Danqi Chen, Eunsol Choi, Michael
  Collins, Kelvin Guu, Hannaneh Hajishirzi, Kenton Lee, Jennimaria Palomaki,
  Colin Raffel, Adam Roberts, Tom Kwiatkowski, Patrick Lewis, Yuxiang Wu,
  Heinrich K\"uttler, Linqing Liu, Pasquale Minervini, Pontus Stenetorp,
  Sebastian Riedel, Sohee Yang, Minjoon Seo, Gautier Izacard, Fabio Petroni,
  Lucas Hosseini, Nicola~De Cao, Edouard Grave, Ikuya Yamada, Sonse Shimaoka,
  Masatoshi Suzuki, Shumpei Miyawaki, Shun Sato, Ryo Takahashi, Jun Suzuki,
  Martin Fajcik, Martin Docekal, Karel Ondrej, Pavel Smrz, Hao Cheng, Yelong
  Shen, Xiaodong Liu, Pengcheng He, Weizhu Chen, Jianfeng Gao, Barlas Oguz,
  Xilun Chen, Vladimir Karpukhin, Stan Peshterliev, Dmytro Okhonko, Michael
  Schlichtkrull, Sonal Gupta, Yashar Mehdad, and Wen-tau Yih. 2021.
\newblock \href {https://proceedings.mlr.press/v133/min21a.html} {{NeurIPS 2020
  EfficientQA Competition: Systems, Analyses and Lessons Learned}}.
\newblock In \emph{Proceedings of the NeurIPS 2020 Competition and
  Demonstration Track}, volume 133 of \emph{Proceedings of Machine Learning
  Research}, pages 86--111. PMLR.

\bibitem[{Mishra and Arunkumar(2021)}]{mishra2021robust}
Swaroop Mishra and Anjana Arunkumar. 2021.
\newblock {How Robust are Model Rankings: A Leaderboard Customization Approach
  for Equitable Evaluation}.
\newblock In \emph{Proceedings of the AAAI Conference on Artificial
  Intelligence}, volume~35, pages 13561--13569.

\bibitem[{Munda(2012)}]{munda}
Giuseppe Munda. 2012.
\newblock \href {http://www.jstor.org/stable/23325434} {Choosing aggregation
  rules for composite indicators}.
\newblock \emph{Social Indicators Research}, 109(3):337--354.

\bibitem[{Nadeem et~al.(2021)Nadeem, Bethke, and
  Reddy}]{nadeem-etal-2021-stereoset}
Moin Nadeem, Anna Bethke, and Siva Reddy. 2021.
\newblock \href {https://doi.org/10.18653/v1/2021.acl-long.416} {{S}tereo{S}et:
  Measuring stereotypical bias in pretrained language models}.
\newblock In \emph{Proceedings of the 59th Annual Meeting of the Association
  for Computational Linguistics and the 11th International Joint Conference on
  Natural Language Processing (Volume 1: Long Papers)}, pages 5356--5371,
  Online. Association for Computational Linguistics.

\bibitem[{Nangia et~al.(2020)Nangia, Vania, Bhalerao, and
  Bowman}]{nangia-etal-2020-crows}
Nikita Nangia, Clara Vania, Rasika Bhalerao, and Samuel~R. Bowman. 2020.
\newblock \href {https://doi.org/10.18653/v1/2020.emnlp-main.154}
  {{C}row{S}-pairs: A challenge dataset for measuring social biases in masked
  language models}.
\newblock In \emph{Proceedings of the 2020 Conference on Empirical Methods in
  Natural Language Processing (EMNLP)}, pages 1953--1967, Online. Association
  for Computational Linguistics.

\bibitem[{Nie{\ss}l et~al.(2022)Nie{\ss}l, Herrmann, Wiedemann, Casalicchio,
  and Boulesteix}]{niessl2022over}
Christina Nie{\ss}l, Moritz Herrmann, Chiara Wiedemann, Giuseppe Casalicchio,
  and Anne-Laure Boulesteix. 2022.
\newblock {Over-optimism in Benchmark Studies and the Multiplicity of Design
  and Analysis Options when Interpreting Their Results}.
\newblock \emph{Wiley Interdisciplinary Reviews: Data Mining and Knowledge
  Discovery}, 12(2):e1441.

\bibitem[{Nurmi(1983)}]{nurmi1983voting}
Hannu Nurmi. 1983.
\newblock Voting procedures: A summary analysis.
\newblock \emph{British Journal of Political Science}, 13(2):181--208.

\bibitem[{Ott et~al.(2022)Ott, Barbosa-Silva, Blagec, Brauner, and
  Samwald}]{ott2022mapping}
Simon Ott, Adriano Barbosa-Silva, Kathrin Blagec, Jan Brauner, and Matthias
  Samwald. 2022.
\newblock {Mapping Global Dynamics of Benchmark Creation and Saturation in
  Artificial Intelligence}.
\newblock \emph{Nature Communications}, 13(1):6793.

\bibitem[{Peyrard et~al.(2017)Peyrard, Botschen, and
  Gurevych}]{peyrard-etal-2017-learning}
Maxime Peyrard, Teresa Botschen, and Iryna Gurevych. 2017.
\newblock \href {https://doi.org/10.18653/v1/W17-4510} {Learning to score
  system summaries for better content selection evaluation.}
\newblock In \emph{Proceedings of the Workshop on New Frontiers in
  Summarization}, pages 74--84, Copenhagen, Denmark. Association for
  Computational Linguistics.

\bibitem[{Procaccia et~al.(2007)Procaccia, Rosenschein, and
  Kaminka}]{procaccia2007robustness}
Ariel~D Procaccia, Jeffrey~S Rosenschein, and Gal~A Kaminka. 2007.
\newblock {On the Robustness of Preference Aggregation in Noisy Environments}.
\newblock In \emph{Proceedings of the 6th international joint conference on
  Autonomous agents and multiagent systems}, pages 1--7.

\bibitem[{Radford et~al.(2019)Radford, Wu, Child, Luan, Amodei, Sutskever
  et~al.}]{radford2019language}
Alec Radford, Jeffrey Wu, Rewon Child, David Luan, Dario Amodei, Ilya
  Sutskever, et~al. 2019.
\newblock {Language Models are Unsupervised Multitask Learners}.
\newblock \emph{OpenAI blog}, 1(8):9.

\bibitem[{Raffel et~al.(2020)Raffel, Shazeer, Roberts, Lee, Narang, Matena,
  Zhou, Li, Liu et~al.}]{raffel2020exploring}
Colin Raffel, Noam Shazeer, Adam Roberts, Katherine Lee, Sharan Narang, Michael
  Matena, Yanqi Zhou, Wei Li, Peter~J Liu, et~al. 2020.
\newblock {Exploring the Limits of Transfer Learning with a Unified
  Text-to-text Transformer}.
\newblock \emph{J. Mach. Learn. Res.}, 21(140):1--67.

\bibitem[{Raji et~al.(2021)Raji, Denton, Bender, Hanna, and
  Paullada}]{raji2021ai}
Inioluwa~Deborah Raji, Emily Denton, Emily~M Bender, Alex Hanna, and
  Amandalynne Paullada. 2021.
\newblock {AI and the Everything in the Whole Wide World Benchmark}.
\newblock In \emph{Thirty-fifth Conference on Neural Information Processing
  Systems Datasets and Benchmarks Track (Round 2)}.

\bibitem[{Rodriguez et~al.(2021)Rodriguez, Barrow, Hoyle, Lalor, Jia, and
  Boyd-Graber}]{rodriguez-etal-2021-evaluation}
Pedro Rodriguez, Joe Barrow, Alexander~Miserlis Hoyle, John~P. Lalor, Robin
  Jia, and Jordan Boyd-Graber. 2021.
\newblock \href {https://doi.org/10.18653/v1/2021.acl-long.346} {Evaluation
  examples are not equally informative: How should that change {NLP}
  leaderboards?}
\newblock In \emph{Proceedings of the 59th Annual Meeting of the Association
  for Computational Linguistics and the 11th International Joint Conference on
  Natural Language Processing (Volume 1: Long Papers)}, pages 4486--4503,
  Online. Association for Computational Linguistics.

\bibitem[{Rogers(2019)}]{rogers2019transformers}
Anna Rogers. 2019.
\newblock {How the Transformers Broke NLP Leaderboards}.
\newblock \url{https://hackingsemantics.xyz/2019/leaderboards}.

\bibitem[{Ruder(2021)}]{ruder2021benchmarking}
Sebastian Ruder. 2021.
\newblock {Challenges and Opportunities in NLP Benchmarking}.
\newblock \url{http://ruder.io/nlp-benchmarking}.

\bibitem[{Salazar et~al.(2020)Salazar, Liang, Nguyen, and
  Kirchhoff}]{salazar-etal-2020-masked}
Julian Salazar, Davis Liang, Toan~Q. Nguyen, and Katrin Kirchhoff. 2020.
\newblock \href {https://doi.org/10.18653/v1/2020.acl-main.240} {Masked
  language model scoring}.
\newblock In \emph{Proceedings of the 58th Annual Meeting of the Association
  for Computational Linguistics}, pages 2699--2712, Online. Association for
  Computational Linguistics.

\bibitem[{Sanh et~al.(2019)Sanh, Debut, Chaumond, and
  Wolf}]{sanh2019distilbert}
Victor Sanh, Lysandre Debut, Julien Chaumond, and Thomas Wolf. 2019.
\newblock {DistilBERT, a distilled version of BERT: smaller, faster, cheaper
  and lighter}.
\newblock \emph{arXiv preprint arXiv:1910.01108}.

\bibitem[{Shavrina and Malykh(2021)}]{shavrina2021not}
Tatiana Shavrina and Valentin Malykh. 2021.
\newblock {How not to lie with a benchmark: rearranging NLP leaderboards}.
\newblock In \emph{{I (Still) Can't Believe It's Not Better! NeurIPS 2021
  Workshop}}.

\bibitem[{Shin et~al.(2021)Shin, Mun, On, Kang, Han, and Kim}]{shin2021winning}
Minchul Shin, Jonghwan Mun, Kyoung-Woon On, Woo-Young Kang, Gunsoo Han, and
  Eun-Sol Kim. 2021.
\newblock {Winning the ICCV'2021 VALUE Challenge: Task-aware Ensemble and
  Transfer Learning with Visual Concepts}.
\newblock \emph{arXiv preprint arXiv:2110.06476}.

\bibitem[{Suppes(1965)}]{suppes1965preference}
P~Suppes. 1965.
\newblock Preference, utility and subjective probability. inhandbook of
  mathematical psychology, ed. rd luce, rr bush and eh galanter, 3, 249--410.

\bibitem[{Varshney et~al.(2022)Varshney, Mishra, and
  Baral}]{varshney-etal-2022-ildae}
Neeraj Varshney, Swaroop Mishra, and Chitta Baral. 2022.
\newblock \href {https://doi.org/10.18653/v1/2022.acl-long.240} {{ILDAE}:
  Instance-level difficulty analysis of evaluation data}.
\newblock In \emph{Proceedings of the 60th Annual Meeting of the Association
  for Computational Linguistics (Volume 1: Long Papers)}, pages 3412--3425,
  Dublin, Ireland. Association for Computational Linguistics.

\bibitem[{Wang et~al.(2019{\natexlab{a}})Wang, Pruksachatkun, Nangia, Singh,
  Michael, Hill, Levy, and Bowman}]{wang2019superglue}
Alex Wang, Yada Pruksachatkun, Nikita Nangia, Amanpreet Singh, Julian Michael,
  Felix Hill, Omer Levy, and Samuel Bowman. 2019{\natexlab{a}}.
\newblock {SuperGLUE: A Stickier Benchmark for General-purpose Language
  Understanding Systems}.
\newblock \emph{Advances in Neural Information Processing Dystems}, 32.

\bibitem[{Wang et~al.(2018)Wang, Singh, Michael, Hill, Levy, and
  Bowman}]{wang-etal-2018-glue}
Alex Wang, Amanpreet Singh, Julian Michael, Felix Hill, Omer Levy, and Samuel
  Bowman. 2018.
\newblock \href {https://doi.org/10.18653/v1/W18-5446} {{GLUE}: A multi-task
  benchmark and analysis platform for natural language understanding}.
\newblock In \emph{Proceedings of the 2018 {EMNLP} Workshop {B}lackbox{NLP}:
  Analyzing and Interpreting Neural Networks for {NLP}}, pages 353--355,
  Brussels, Belgium. Association for Computational Linguistics.

\bibitem[{Wang et~al.(2021)Wang, Xu, Wang, Wang, Gan, Cheng, Gao, Awadallah,
  and Li}]{advglue}
Boxin Wang, Chejian Xu, Shuohang Wang, Shuohang Wang, Zhe Gan, Yu~Cheng,
  Jianfeng Gao, Ahmed Awadallah, and Bo~Li. 2021.
\newblock \href
  {https://datasets-benchmarks-proceedings.neurips.cc/paper/2021/file/335f5352088d7d9bf74191e006d8e24c-Paper-round2.pdf}
  {{Adversarial GLUE: A Multi-Task Benchmark for Robustness Evaluation of
  Language Models}}.
\newblock In \emph{Proceedings of the Neural Information Processing Systems
  Track on Datasets and Benchmarks}, volume~1.

\bibitem[{Wang et~al.(2019{\natexlab{b}})Wang, Bi, Yan, Wu, Xia, Bao, Peng, and
  Si}]{wang2019structbert}
Wei Wang, Bin Bi, Ming Yan, Chen Wu, Jiangnan Xia, Zuyi Bao, Liwei Peng, and
  Luo Si. 2019{\natexlab{b}}.
\newblock {StructBERT: Incorporating Language Structures into Pre-training for
  Deep Language Understanding}.
\newblock In \emph{International Conference on Learning Representations}.

\bibitem[{Waseem et~al.(2021)Waseem, Lulz, Bingel, and
  Augenstein}]{waseem2021disembodied}
Zeerak Waseem, Smarika Lulz, Joachim Bingel, and Isabelle Augenstein. 2021.
\newblock {Disembodied Machine Learning: On the Illusion of Objectivity in
  NLP}.
\newblock \emph{arXiv preprint arXiv:2101.11974}.

\bibitem[{Webb(2000)}]{webb2000multiboosting}
Geoffrey~I Webb. 2000.
\newblock {MultiBoosting: A Technique for Combining Boosting and Wagging}.
\newblock \emph{Machine learning}, 40(2):159--196.

\bibitem[{Williams et~al.(2018)Williams, Nangia, and
  Bowman}]{williams-etal-2018-broad}
Adina Williams, Nikita Nangia, and Samuel Bowman. 2018.
\newblock \href {https://doi.org/10.18653/v1/N18-1101} {A broad-coverage
  challenge corpus for sentence understanding through inference}.
\newblock In \emph{Proceedings of the 2018 Conference of the North {A}merican
  Chapter of the Association for Computational Linguistics: Human Language
  Technologies, Volume 1 (Long Papers)}, pages 1112--1122, New Orleans,
  Louisiana. Association for Computational Linguistics.

\bibitem[{Wolf et~al.(2020)Wolf, Debut, Sanh, Chaumond, Delangue, Moi, Cistac,
  Rault, Louf, Funtowicz, Davison, Shleifer, von Platen, Ma, Jernite, Plu, Xu,
  Le~Scao, Gugger, Drame, Lhoest, and Rush}]{wolf-etal-2020-transformers}
Thomas Wolf, Lysandre Debut, Victor Sanh, Julien Chaumond, Clement Delangue,
  Anthony Moi, Pierric Cistac, Tim Rault, Remi Louf, Morgan Funtowicz, Joe
  Davison, Sam Shleifer, Patrick von Platen, Clara Ma, Yacine Jernite, Julien
  Plu, Canwen Xu, Teven Le~Scao, Sylvain Gugger, Mariama Drame, Quentin Lhoest,
  and Alexander Rush. 2020.
\newblock \href {https://doi.org/10.18653/v1/2020.emnlp-demos.6} {Transformers:
  State-of-the-art natural language processing}.
\newblock In \emph{Proceedings of the 2020 Conference on Empirical Methods in
  Natural Language Processing: System Demonstrations}, pages 38--45, Online.
  Association for Computational Linguistics.

\bibitem[{Zhang et~al.(2019)Zhang, Han, Liu, Jiang, Sun, and
  Liu}]{zhang-etal-2019-ernie}
Zhengyan Zhang, Xu~Han, Zhiyuan Liu, Xin Jiang, Maosong Sun, and Qun Liu. 2019.
\newblock \href {https://doi.org/10.18653/v1/P19-1139} {{ERNIE}: Enhanced
  language representation with informative entities}.
\newblock In \emph{Proceedings of the 57th Annual Meeting of the Association
  for Computational Linguistics}, pages 1441--1451, Florence, Italy.
  Association for Computational Linguistics.

\bibitem[{Zhao et~al.(2018)Zhao, Wang, Yatskar, Ordonez, and
  Chang}]{zhao-etal-2018-gender}
Jieyu Zhao, Tianlu Wang, Mark Yatskar, Vicente Ordonez, and Kai-Wei Chang.
  2018.
\newblock \href {https://doi.org/10.18653/v1/N18-2003} {Gender bias in
  coreference resolution: Evaluation and debiasing methods}.
\newblock In \emph{Proceedings of the 2018 Conference of the North {A}merican
  Chapter of the Association for Computational Linguistics: Human Language
  Technologies, Volume 2 (Short Papers)}, pages 15--20, New Orleans, Louisiana.
  Association for Computational Linguistics.

\bibitem[{Zhou et~al.(2021)Zhou, Chen, Jin, and Wang}]{zhou-etal-2021-hulk}
Xiyou Zhou, Zhiyu Chen, Xiaoyong Jin, and William~Yang Wang. 2021.
\newblock \href {https://doi.org/10.18653/v1/2021.eacl-demos.39} {{HULK}: An
  energy efficiency benchmark platform for responsible natural language
  processing}.
\newblock In \emph{Proceedings of the 16th Conference of the European Chapter
  of the Association for Computational Linguistics: System Demonstrations},
  pages 329--336, Online. Association for Computational Linguistics.

\end{thebibliography}
